\documentclass[11pt,letterpaper]{article}
\usepackage[margin=1in]{geometry}
\usepackage[T1]{fontenc}
\usepackage{times}
\usepackage{amsmath,amssymb,amsthm,mathtools,bm}
\usepackage{booktabs,array,longtable,graphicx}
\usepackage{aliascnt,enumitem,microtype,xcolor}
\usepackage{algorithm,algpseudocode}
\usepackage[numbers,sort&compress]{natbib}
\usepackage[hidelinks]{hyperref}
\hypersetup{pdftitle={Self-Normalized Inference from Riemannian Stochastic Gradient Paths},pdfauthor={Xiaotian Chang, Qirui Hu, Yi Liu}}
\usepackage[nameinlink,capitalize,noabbrev]{cleveref}

\newtheorem{theorem}{Theorem}[section]
\newaliascnt{proposition}{theorem}
\newtheorem{proposition}[proposition]{Proposition}
\aliascntresetthe{proposition}
\newaliascnt{corollary}{theorem}
\newtheorem{corollary}[corollary]{Corollary}
\aliascntresetthe{corollary}
\newaliascnt{lemma}{theorem}
\newtheorem{lemma}[lemma]{Lemma}
\aliascntresetthe{lemma}
\newaliascnt{assumption}{theorem}
\newtheorem{assumption}[assumption]{Assumption}
\aliascntresetthe{assumption}
\theoremstyle{definition}
\newaliascnt{definition}{theorem}

\aliascntresetthe{definition}
\newaliascnt{example}{theorem}

\aliascntresetthe{example}
\theoremstyle{remark}
\newaliascnt{remark}{theorem}
\newtheorem{remark}[remark]{Remark}
\aliascntresetthe{remark}

\crefname{theorem}{Theorem}{Theorems}
\crefname{proposition}{Proposition}{Propositions}
\crefname{corollary}{Corollary}{Corollaries}
\crefname{lemma}{Lemma}{Lemmas}
\crefname{assumption}{Assumption}{Assumptions}
\crefname{definition}{Definition}{Definitions}
\crefname{example}{Example}{Examples}
\crefname{remark}{Remark}{Remarks}
\crefname{algorithm}{Algorithm}{Algorithms}

\newcommand{\M}{\mathcal M}
\newcommand{\F}{\mathcal F}
\newcommand{\R}{\mathbb R}
\newcommand{\E}{\mathbb E}
\newcommand{\Pp}{\mathbb P}
\newcommand{\cN}{\mathcal N}
\newcommand{\cT}{\mathcal T}
\newcommand{\cC}{\mathcal C}
\newcommand{\grad}{\operatorname{grad}}
\newcommand{\Hess}{\operatorname{Hess}}
\newcommand{\Exp}{\operatorname{Exp}}
\newcommand{\Log}{\operatorname{Log}}
\newcommand{\dist}{\operatorname{dist}}
\newcommand{\tr}{\operatorname{tr}}
\newcommand{\vecop}{\operatorname{vec}}
\newcommand{\Id}{\mathrm I}
\newcommand{\op}{\mathrm{op}}
\newcommand{\ind}{\mathbf 1}
\newcommand{\ip}[2]{\left\langle #1,#2\right\rangle}
\newcommand{\norm}[1]{\left\lVert #1\right\rVert}
\newcommand{\abs}[1]{\left\lvert #1\right\rvert}
\newcommand{\ot}{\mathbin{\otimes}}
\newcommand{\transpose}{\mathsf T}
\newcommand{\rightsquig}{\rightsquigarrow}

\title{Self-Normalized Inference from Riemannian Stochastic Gradient Paths}
\author{%
Xiaotian Chang\thanks{Division of Mathematical Sciences, Nanyang Technological University.}\quad
Qirui Hu\thanks{School of Statistics and Data Science, Shanghai University of Finance and Economics.}\quad
Yi Liu\thanks{York University. Corresponding author. Email: \texttt{yliu42@yorku.ca}.}%
}
\date{}

\begin{document}
\maketitle
\begin{abstract}
Riemannian stochastic gradient paths estimate constrained parameters, but confidence regions usually require estimating the Hessian and gradient-noise covariance. We develop self-normalized inference from one path by expressing retained iterates in a common random tangent space. Their average and centered partial sums share an unknown covariance factor that cancels in a Brownian bridge pivot. Under suitable regularity conditions, a joint expansion of the averaged estimator and integrated path controls nonlinear coordinate errors using conditional second moments. The procedure uses fixed-dimensional memory and extends to nondegenerate smooth functionals and alternative local coordinates. Numerical studies show competitive coverage relative to plug-in Wald inference while avoiding Hessian and residual-density estimation.
\end{abstract}

\section{Introduction}\label{sec:intro}

Many statistical and machine learning parameters have geometric constraints: directions lie on spheres, covariance matrices and diffusion tensors in the symmetric positive definite (SPD) matrix cone, and principal subspaces on a Grassmann manifold. These lead to population problems
\begin{equation}\label{eq:population-problem}
 x_\star\in\arg\min_{x\in\M}F(x),
 \qquad
 F(x)=\E\{\ell(x;Z)\},
\end{equation}
where $(\M,g)$ is a finite-dimensional Riemannian manifold. With sequential observations or a single sample pass, Riemannian stochastic gradients naturally estimate \eqref{eq:population-problem} \citep{Bonnabel2013,Tripuraneni2018,Durmus2021}. Their optimization and averaging theory is well developed \citep{Absil2008,Boumal2023}.

Inference is less direct. For an unbiased stochastic Riemannian gradient $G(x,Z)$, write in the tangent space at a regular minimizer, using $\ot$ for tangent outer products,
\begin{equation}\label{eq:sandwich-intro}
\begin{gathered}
 H=\Hess F(x_\star),\quad \Omega=H^{-1}\Sigma H^{-1},\\
 \Sigma=\E\{G(x_\star,Z)\ot G(x_\star,Z)\}.
\end{gathered}
\end{equation}
Averaged stochastic approximation has covariance \eqref{eq:sandwich-intro} \citep{PolyakJuditsky1992,Tripuraneni2018}. Plug-in Wald inference estimates $H$ and $\Sigma$ in a common tangent space. This adds a second-order calculation to a first-order optimizer. For affine-invariant matrix means, it involves differentiating a matrix logarithm; for conditional-median sensing, the population Hessian contains an unknown residual density. Even an analytic observation Hessian can be difficult to average accurately: in multi-frequency circular estimation its magnitude grows with frequency while the stochastic gradient remains bounded. A path-based procedure avoids these additional Hessian calculations and their estimation variability.

Self-normalization uses the same partial sum process in numerator and denominator, cancelling the unknown linear covariance factor in Euclidean time series and stochastic approximation \citep{Shao2010,ShaoCorrigendum2010,Lee2022}. On a manifold, constructing such sums requires common coordinates and control of nonlinear coordinate errors accumulated along the gradient path.

Classical stochastic approximation and Polyak--Ruppert averaging provide the underlying estimation theory \citep{PolyakJuditsky1992,KushnerYin2003}. Riemannian results establish convergence, averaging, streaming $k$-PCA, and fixed-step approximation \citep{Bonnabel2013,Tripuraneni2018,Durmus2021}. Euclidean online inference uses covariance estimates and batch means \citep{Chen2020}, perturbed trajectories \citep{FangXuYang2018}, or HiGrad branching \citep{SuZhu2023}. Random scaling instead uses a single path \citep{Lee2022,Li2022}, including stochastic subgradient quantile regression where it avoids residual-density estimation \citep{Lee2025QR}. Our construction reduces exactly to Euclidean random scaling; the additional difficulty is preserving its common covariance factor through nonlinear coordinates and a random tangent anchor.

Inference from manifold observations includes intrinsic and extrinsic mean limits \citep{BhattacharyaPatrangenaru2005}, distributions meeting the cut locus \citep{HotzLeWood2024}, and empirical likelihood for Fr\'echet means \citep{KurisuOtsu2026}. Here the inferential input is the dependent trajectory of an online optimizer, whose covariance is determined by stochastic gradients.

To construct inference from a single path, we express retained iterates in the tangent space at a burn-in iterate and accumulate their average and centered partial sums in one pass. The central difficulty is to show that nonlinear coordinates preserve the common covariance factor needed for self-normalization. We address this through a joint Polyak--Ruppert expansion of the averaged estimator and integrated path, controlling the nonlinear remainders with conditional second moments under the localization, stability, and martingale conditions in Section~\ref{sec:assumptions}. This expansion yields a Brownian bridge pivot depending only on dimension, and hence confidence regions without Hessian or gradient-noise covariance estimation. Because the same cancellation survives smooth projection and changes of local coordinates, the construction also provides inference for interpretable contrasts such as squared principal-component loadings.

The rest of the paper is organized as follows. Section~\ref{sec:method} gives the methodology, Section~\ref{sec:theory} presents the theoretical results, Section~\ref{sec:simulations} reports simulations and a data illustration, and Section~\ref{sec:conclusion} discusses implications. The appendices contain additional experiments, geometric specializations, technical lemmas, and proofs.

\section{Methodology}\label{sec:method}

Let $(\M,g)$ be a connected $d$-dimensional Riemannian manifold, with tangent inner product $\ip{\cdot}{\cdot}_x$ and norm $\norm{\cdot}_x$. We write $\Id$ for the identity operator, $\norm{\cdot}_{\op}$ for the operator norm, and $\vecop$ for column-stacking of a matrix. On a strongly convex normal neighborhood, $\Exp_x$ is the exponential map, $\Log_x$ its local inverse, and $\mathsf P_{x\to y}$ parallel transport along the minimizing geodesic.

\subsection{Recursion and averaging in a fixed chart}
Let $\F_k=\sigma(Z_1,\ldots,Z_k,x_0)$ and let $G_{k+1}$ be an adapted unbiased gradient field:
$\E\{G_{k+1}(x)\mid\F_k\}=\grad F(x)$. For i.i.d. observations, $G_{k+1}(x)=G(x,Z_{k+1})$. We use
\begin{equation}\label{eq:rsgd}
 x_{k+1}=\mathcal R_{x_k}\{-\eta_{k+1}G_{k+1}(x_k)\},
 \quad \eta_k=\eta_0k^{-\beta},
\end{equation}
where $\eta_0>0$ and $\mathcal R_x$ is a $C^2$ retraction, with $\mathcal R_x(0)=x$ and $D\mathcal R_x(0)=\Id$. The exponential map $\mathcal R_x=\Exp_x$ is one such choice. Discard the first $b_n$ iterates, where the deterministic burn-in satisfies
\begin{equation}\label{eq:burnin}
 b_n\to\infty,\quad b_n/n\to0,\quad N_n=n-b_n,
\end{equation}
Thus the number of discarded iterates grows with $n$, while their fraction vanishes and $N_n/n\to1$. This is the diverging burn-in used below. Set $a_n=x_{b_n}$ to fix the tangent space for averaging; if the recursion converges to $x_\star$, this anchor is consistent. Assumptions~\ref{ass:geometry} and \ref{ass:stability} specify the localization and stability needed for the retained path. On the common normal neighborhood define
\begin{equation}\label{eq:vk}
 v_{k,n}=\Log_{a_n}(x_k)\in T_{a_n}\M,\quad b_n<k\le n,
\end{equation}
and the fixed-chart Polyak--Ruppert estimate
{
\begin{equation*}
 \bar v_n=N_n^{-1}\sum_{k=b_n+1}^nv_{k,n},\quad
 \widehat x_n=\Exp_{a_n}(\bar v_n).
\end{equation*}
}
Smooth inverse retractions provide alternative coordinates; \Cref{prop:coordinate-reduction} establishes their first order equivalence.

\subsection{Centered path and confidence region}
For $0\le m\le N_n$, with $S_{0,n}=C_{0,n}=0$, define
{
\begin{align*}
 S_{m,n}&=\sum_{j=1}^mv_{b_n+j,n},\\
 C_{m,n}&=S_{m,n}-\frac{m}{N_n}S_{N_n,n}.
\end{align*}
}The centered sums give
\begin{equation}\label{eq:self-normalizer}
 \widehat V_n=N_n^{-2}\sum_{m=1}^{N_n}C_{m,n}\ot C_{m,n},
\end{equation}
where $(u\ot v)w=u\ip{v}{w}_{a_n}$. Since the centered sums fluctuate on the $\sqrt{N_n}$ scale, this normalization gives a nondegenerate limit. For a candidate $x$, define $e_n(x)=\bar v_n-\Log_{a_n}(x)$ in orthonormal tangent coordinates and put
{
\begin{equation}\label{eq:test-statistic}
 T_n(x)=N_n e_n(x)^\transpose\widehat V_n^{-1}e_n(x).
\end{equation}
}
Let $W_d$ be a standard $d$-dimensional Brownian motion on $[0,1]$ and define
\begin{equation}\label{eq:pivot-limit}
\begin{gathered}
 B_d(r)=W_d(r)-rW_d(1),\\
 Q_d=\int_0^1B_d(r)B_d(r)^\transpose\,dr,\\
 \cT_d=W_d(1)^\transpose Q_d^{-1}W_d(1).
\end{gathered}
\end{equation}
For $0<\alpha<1$, let $c_{d,1-\alpha}$ be the $(1-\alpha)$ quantile of $\cT_d$ and report
\begin{equation}\label{eq:confidence-region}
 \cC_{n,1-\alpha}=\{x:T_n(x)\le c_{d,1-\alpha}\}
\end{equation}
within the chart domain.

In tangent coordinates, \eqref{eq:confidence-region} is an ellipsoid centered at $\bar v_n$. An eigenvalue $\widehat\lambda_j$ of $\widehat V_n$ gives semiaxis $\sqrt{c_{d,1-\alpha}\widehat\lambda_j/N_n}$; the exponential map sends this ellipsoid to the manifold.

\subsection{Computation in one pass}\label{sec:algorithm}
Write $N=N_n$, $S_m=S_{m,n}$, and maintain
$A_N=\sum_{m=1}^N S_m\ot S_m$ and $B_N=\sum_{m=1}^N mS_m$. Expanding the centered products gives
\begin{equation}\label{eq:online-V}
\begin{split}
 \widehat V_n=\frac1{N^2}\bigg[
 A_N-\frac{B_N\ot S_N+S_N\ot B_N}{N}\\
 {}+\frac{q_N}{N^2}S_N\ot S_N\bigg],
 \quad q_N=\frac{N(N+1)(2N+1)}6.
\end{split}
\end{equation}

\begin{algorithm}[tb]
\caption{Estimation and inference from a single path}\label{alg:one-path}
{
\begin{algorithmic}[1]
\Require Initial point $x_0$, gradient stream, gains $\eta_k$, sample size $n$, burn-in $b_n$, and critical value $c_{d,1-\alpha}$.
\Ensure Estimate $\widehat x_n$ and region $\cC_{n,1-\alpha}$.
\State Run \eqref{eq:rsgd} for $b_n$ steps.
\State Set $a_n\gets x_{b_n}$ and $N\gets n-b_n$.
\State Initialize $S=B=0\in T_{a_n}\M$ and $A=0$.
\For{$m=1,\ldots,N$}
  \State Continue \eqref{eq:rsgd} to obtain $x_{b_n+m}$.
  \State $v_m\gets\Log_{a_n}(x_{b_n+m})$.
  \State $S\gets S+v_m$.
  \State $A\gets A+S\ot S$.
  \State $B\gets B+mS$.
\EndFor
\State $\bar v_n\gets S/N$ and $\widehat x_n\gets\Exp_{a_n}(\bar v_n)$.
\State Compute $\widehat V_n$ by \eqref{eq:online-V}.
\State \Return $\widehat x_n$ and $\cC_{n,1-\alpha}$ from \eqref{eq:confidence-region}.
\end{algorithmic}
}
\end{algorithm}
Each update uses $O(d^2)$ arithmetic apart from evaluating the chart, and the accumulators use $O(d^2)$ memory. Full-region inference adds a final matrix solve. Equation~\eqref{eq:online-V} avoids storing or replaying the trajectory. A vanishing nonnegative ridge preserves the pivot under nondegeneracy (\Cref{prop:ridge-stability}).

\label{sec:calibration-coordinates}
The Brownian critical values are shared across data sets of the same tangent dimension. The bridge has the Karhunen--Lo\`eve representation
\[
 Q_d=\sum_{j=1}^{\infty}\frac{z_jz_j^\transpose}{(\pi j)^2},
 \qquad \cT_d=e^\transpose Q_d^{-1}e,
\]
where $e,z_1,z_2,\ldots$ are independent standard $d$-dimensional Gaussian vectors. Nested truncations with at least $d$ modes are almost surely positive definite; their quadratic forms decrease to $\cT_d$. 

For a truncation after $J$ modes, the expected trace of the omitted bridge Gram terms is at most $d/(\pi^2J)$. This bounds the matrix tail. Nested truncation gives one-sided approximation of the pivot.
The scalar experiments use the critical value $c_{1,0.95}$, displayed as $46.14$. Appendix~\ref{sec:calibration} gives the simulation counts, truncation levels, and full-region critical values.

The tangent basis is arbitrary. For an orthogonal matrix $O$, the estimation error and normalizer become $Oe_n(x)$ and $O\widehat V_nO^\transpose$, leaving \eqref{eq:test-statistic} unchanged. More generally, a jointly smooth family of uniformly $C^2$ local charts with uniformly nonsingular derivatives preserves the first order limit (\Cref{prop:coordinate-reduction}). This permits inverse-retraction or graph coordinates when they are convenient. In Euclidean space, $v_{k,n}=x_k-a_n$: centering removes the common anchor, $\widehat x_n$ is the arithmetic average, and \Cref{alg:one-path} reduces exactly to the random scaling procedure of \citet{Lee2022}.

Wald inference estimates $\Omega=H^{-1}\Sigma H^{-1}$ and uses chi-square critical values. Self-normalization retains the random path shape $LQ_dL^\transpose$, where $L=H^{-1}\Sigma^{1/2}$, and cancels $L$ by combining numerator and denominator. Both procedures have root-$N_n$ radii under their respective assumptions; \Cref{prop:wald-comparison} gives their volume limits. The computational distinction is estimated Hessian and noise covariance versus centered trajectory accumulation.

\section{Theoretical Results}\label{sec:theory}
\subsection{Assumptions}\label{sec:assumptions}
The theory studies inference at an isolated local minimizer $x_\star$. Fix orthonormal coordinates on $T_{x_\star}\M$ and write $\Delta_k=\Log_{x_\star}(x_k)$ on the local chart, with an arbitrary measurable extension outside it. The following conditions identify the geometric and stochastic inputs; sufficient conditions for verifying them are given in \Cref{sec:assumption-discussion}.

\begin{assumption}[Local geometry and localization]\label{ass:geometry}
There are strongly convex normal neighborhoods $U_0,U_1$ with $x_\star\in U_0$, $\overline U_0\subset U_1$, and $\overline U_1$ compact. Logarithms and parallel transport are smooth on the relevant pairs. For some $r_{\mathcal R}>0$, the retraction is $C^2$ on $\{(x,v):x\in\overline U_0,\ \norm v_x\le r_{\mathcal R}\}$ and maps this tube into $U_1$. For $b_n$ in \eqref{eq:burnin},
\[
 \Pp\{x_k\in U_0\text{ for all }b_n\le k\le n\}\to1.
\]
\end{assumption}

\begin{assumption}[Regular local mean field]\label{ass:objective}
$\grad F(x_\star)=0$, and $H=\Hess F(x_\star)$ is self-adjoint and positive definite. For some $\rho\in(0,1]$ and $L_F<\infty$,
\begin{equation}\label{eq:mean-field-holder}
\begin{split}
 &\norm{\mathsf P_{x\to x_\star}\grad F(x)-H\Log_{x_\star}(x)}\\
 &\qquad\le L_F\dist(x,x_\star)^{1+\rho},\quad x\in U_1.
\end{split}
\end{equation}
\end{assumption}

\begin{assumption}[Stochastic gradient]\label{ass:oracle}
The gradient field is jointly $\F_{k+1}\otimes\mathcal B(U_1)$-measurable in local trivializations and conditionally unbiased. Put
$\epsilon_{k+1}(x)=G_{k+1}(x)-\grad F(x)$ and $\xi_{k+1}=\epsilon_{k+1}(x_\star)$. For $C_G<\infty$ and $\nu>0$,
\begin{align}
 \E\{\norm{G_{k+1}(x)}_x^2\mid\F_k\}&\le C_G,\label{eq:gradient-second-moment}\\
 \E\big[\norm{\mathsf P_{x\to x_\star}\epsilon_{k+1}(x)-\xi_{k+1}}^2
 \mid\F_k\big]&\le C_G\dist(x,x_\star)^{2\nu}.\label{eq:noise-continuity}
\end{align}
These assertions also hold at $\F_k$-measurable $U_1$-valued random arguments.
\end{assumption}

\begin{assumption}[Step size and stability in mean square]\label{ass:stability}
The gain exponent satisfies
\begin{equation}\label{eq:beta-range}
 \max\{1/2,1/(1+\rho)\}<\beta<1.
\end{equation}
For a fixed smooth increment cutoff as in \Cref{lem:local-recursion}, independent of $n$, let $\Delta_{k,n}^{\rm loc}$ be the burn-in-reset auxiliary process defined in \Cref{app:localization}, \eqref{eq:global-local-recursion}. It agrees with $\Delta_k$ on the retained localization and small-increment event. Some $C_\Delta<\infty$ and $n_0\in\mathbb N$ satisfy
\begin{equation}\label{eq:l2-rate}
 \sup_{n\ge n_0}\sup_{b_n\le k\le n}
 \frac{\E\norm{\Delta_{k,n}^{\rm loc}}^2}{\eta_k}
 \le C_\Delta.
\end{equation}
\end{assumption}

\begin{assumption}[Martingale invariance principle]\label{ass:fclt}
$(\xi_{k+1},\F_{k+1})$ is a martingale difference sequence. Write $\Sigma_{k+1}=\E(\xi_{k+1}\ot\xi_{k+1}\mid\F_k)$ for its conditional covariance. For a nonnegative definite $\Sigma$,
{
\begin{equation}\label{eq:predictable-covariance}
 \sup_{r\in[0,1]}\left\|
 N_n^{-1}\sum_{k=b_n}^{b_n+\lfloor N_nr\rfloor-1}
 \Sigma_{k+1}-r\Sigma\right\|_{\op}\xrightarrow{p}0.
\end{equation}
}

For each $\varepsilon>0$, the conditional Lindeberg condition holds:
\begin{equation}\label{eq:lindeberg}
\begin{split}
 N_n^{-1}\sum_{k=b_n}^{n-1}
 \E\big[\norm{\xi_{k+1}}^2
 \ind\{\norm{\xi_{k+1}}>\varepsilon\sqrt{N_n}\}\mid\F_k\big]
 \xrightarrow{p}0.
\end{split}
\end{equation}
\end{assumption}

The requirement $\beta(1+\rho)>1$ controls accumulated mean field errors; $\beta>1/2$ controls curvature and retraction errors; $\beta<1$ controls the averaging boundary. For a locally Lipschitz Hessian, $\rho=1$, so the full range $1/2<\beta<1$ is available. Lower H\"older smoothness requires a more rapidly decreasing gain; for example, $\rho=1/2$ gives $2/3<\beta<1$.

These assumptions can be checked separately for the model and the recursion. Almost sure convergence to $x_\star$ gives retained localization for any diverging burn-in. For a smooth i.i.d. loss, the transported Hessian condition
\begin{equation}\label{eq:hessian-holder}
\begin{split}
 &\norm{\mathsf P_{x\to x_\star}\Hess F(x)\mathsf P_{x_\star\to x}-H}_{\op}\\
 &\qquad\le L\dist(x,x_\star)^\rho.
\end{split}
\end{equation}
implies the mean-field expansion. A sufficient condition for noise continuity with $\nu=1$ is the square-integrable local Hessian envelope
\begin{equation}\label{eq:stochastic-hessian-envelope}
 \E\left[\sup_{x\in U_1}\norm{\Hess_x\ell(x;Z)}_{\op}^2\right]<\infty.
\end{equation}
Finite second moments of the true-point gradient then give the martingale covariance and Lindeberg conditions. These checks identify regularity of the statistical model; convergence to the selected minimizer and uniform mean square stability are properties of the stochastic recursion.

For stability, positive definiteness of $H$ and the mean-field expansion give, on a sufficiently small neighborhood,
\begin{equation}\label{eq:local-monotonicity}
\begin{gathered}
 \ip{\Delta}{\mathsf P_{x\to x_\star}\grad F(x)}_{x_\star}
 \ge \mu_0\norm{\Delta}^2,\\
 \Delta=\Log_{x_\star}(x),\qquad \mu_0>0.
\end{gathered}
\end{equation}
Together with conditional second moments, this yields the one-step drift bound \eqref{eq:one-step-drift} for the auxiliary update. Proposition~\ref{prop:stability} converts that drift and a uniform $O(\eta_{b_n})$ initial second moment into \eqref{eq:l2-rate}. Consistency of the anchor alone does not imply this initial moment bound. The distinction matters because the retained path starts at a random iterate: localization keeps it in a valid chart, whereas the moment rate controls the size of the errors accumulated in that chart. The uniformity in $n$ ensures that the same bound remains valid as the burn-in point changes. Once initialized at the required rate, the drift contracts the existing error and the $O(\eta_{k+1}^2)$ term accounts for fresh gradient noise. Appendix~\ref{sec:assumption-discussion} gives the verification details and comparison with existing averaging assumptions.

\begin{proposition}[A drift bound for one update implies the stability rate]\label{prop:stability}
Suppose there is $C_b<\infty$ such that, uniformly for all sufficiently large $n$,
\[
 \E\norm{\Delta_{b_n,n}^{\rm loc}}^2\le C_b\eta_{b_n},
\]
and, for $b_n\le k<n$,
\begin{equation}\label{eq:one-step-drift}
 \E(\norm{\Delta_{k+1,n}^{\rm loc}}^2\mid\F_k)
 \le (1-c_0\eta_{k+1})\norm{\Delta_{k,n}^{\rm loc}}^2+C_0\eta_{k+1}^2
\end{equation}
for constants $c_0,C_0>0$ independent of $n$ and $k$. If $0<\beta<1$, then \eqref{eq:l2-rate} holds. A sufficient condition for \eqref{eq:one-step-drift}, for all sufficiently large $n$, is Assumptions \ref{ass:geometry}--\ref{ass:oracle} together with \eqref{eq:local-monotonicity} on $\overline U_0$.
\end{proposition}

\subsection{Partial sum limit}\label{sec:partial-sum}
By \Cref{lem:localization-transfer}, the measurable extension of $\Delta_k$ does not affect the limits below. Define the retained partial sum process
\begin{equation}\label{eq:delta-process}
 \mathbb D_n(r)
 =\frac{1}{\sqrt{N_n}}
 \sum_{k=b_n+1}^{b_n+\lfloor N_nr\rfloor}\Delta_k,
 \qquad 0\le r\le1.
\end{equation}
The process in \eqref{eq:delta-process} is viewed as a step function in $L^2([0,1];\R^d)$. The endpoint is included separately because evaluation at $r=1$ is not continuous in the $L^2$ topology. This choice matches the statistic: the numerator uses the endpoint, whereas the denominator integrates outer products of the centered path. Joint convergence of these two objects is therefore sufficient for inference. In particular, $L^2$ convergence controls the integrated quadratic form by the Cauchy--Schwarz inequality without requiring uniform convergence of the entire trajectory.

\begin{theorem}\label{thm:integrated-fclt}
Under Assumptions \ref{ass:geometry}--\ref{ass:fclt},
\begin{equation}\label{eq:joint-L2-limit}
\begin{gathered}
 \{\mathbb D_n(\cdot),\mathbb D_n(1)\}\\
 \rightsquigarrow
 \{-H^{-1}\Sigma^{1/2}W_d(\cdot),
   -H^{-1}\Sigma^{1/2}W_d(1)\}
\end{gathered}
\end{equation}
in $L^2([0,1];\R^d)\times\R^d$, where $W_d$ is a standard $d$-dimensional Brownian motion.
\end{theorem}

The endpoint gives the usual averaged-estimator limit, while the $L^2$ component determines the limit of the integrated path products in the normalizer. The proof separates boundary, changing-gain, mean-field, noise, and retraction errors. Their control explains the gain restrictions: squared boundary errors average to order $n^{\beta-1}$, the accumulated mean-field error is of order $n^{1/2-\beta(1+\rho)/2}$, and quadratic geometric errors are of order $n^{1/2-\beta}$. Technical lemmas appear in Appendix~\ref{app:technical}; the proof is in Appendix~\ref{app:partial-sum}.

\subsection{Self-normalized inference}\label{sec:selfnorm-theory}

The endpoint and centered path share $L=H^{-1}\Sigma^{1/2}$. Their limits are $-LW_d(1)$ and $LQ_dL^\transpose$, so an invertible $L$ cancels in the quadratic statistic. The bridge Gram matrix $Q_d$ is almost surely positive definite and independent of $W_d(1)$, giving the continuous pivotal law in \eqref{eq:pivot-limit}. The normalizer thus remains random in the limit. Its role is to carry the same unknown linear transformation as the estimation error, so that inversion removes that transformation. This explains why the critical value comes from a Brownian bridge rather than a chi-square law, even though the averaged estimator itself is asymptotically Gaussian.

\begin{theorem}[Self-normalized Riemannian inference]\label{thm:selfnorm}
Suppose Assumptions \ref{ass:geometry}--\ref{ass:fclt} hold. Let $a_n=x_{b_n}$ and construct $\widehat x_n$ and $\widehat V_n$ from \eqref{eq:vk}--\eqref{eq:self-normalizer}. The local operations are well-defined with probability tending to one (\Cref{lem:chart-domain}). Then:
\begin{enumerate}[label=(\alph*),leftmargin=2.1em]
\item The fixed-chart average has the standard Polyak--Ruppert limit
\begin{equation}\label{eq:intrinsic-clt}
 \sqrt{N_n}\Log_{x_\star}(\widehat x_n)
 \rightsquig
 \cN\{0,H^{-1}\Sigma H^{-1}\}.
\end{equation}
\item With
\[
 \widetilde V_n
 =\mathsf P_{a_n\to x_\star}\widehat V_n\mathsf P_{x_\star\to a_n},
\]
we have
\begin{equation}\label{eq:V-limit}
 \widetilde V_n
 \rightsquig
 H^{-1}\Sigma^{1/2}Q_d\Sigma^{1/2}H^{-1}.
\end{equation}
\item If $\Sigma$ is positive definite, $\widehat V_n$ is invertible with probability tending to one and, at the true parameter,
\begin{equation}\label{eq:pivot-convergence}
 T_n(x_\star)\rightsquig\cT_d.
\end{equation}
Consequently, for $c_{d,1-\alpha}$ the $(1-\alpha)$ quantile of $\cT_d$,
\begin{equation}\label{eq:coverage}
 \Pp\{x_\star\in\cC_{n,1-\alpha}\}\longrightarrow1-\alpha.
\end{equation}
\end{enumerate}
\end{theorem}

The same cancellation holds in the practical chart at $a_n$. The anchor is computed from the observed path, so it need not be independent of the retained iterates. Uniform local chart expansions handle this dependence by applying the same remainder bound throughout the localization neighborhood, including at the random anchor. Its common offset disappears from every centered sum, while the remaining coordinate error is quadratic in the local iterate error. Mean square stability and $\beta>1/2$ make its accumulated contribution negligible at the $\sqrt{N_n}$ scale. Appendix~\ref{app:local} gives the chart expansion, and Appendix~\ref{app:selfnorm} proves the joint bridge limit and \Cref{thm:selfnorm}.

\subsection{Inference for smooth functionals}
For a reported contrast, only its projected covariance needs to be nondegenerate.

\begin{corollary}[Smooth functionals]\label{cor:functionals}
Under Assumptions \ref{ass:geometry}--\ref{ass:fclt}, let $h:\M\to\R^q$ be continuously differentiable near $x_\star$ and set
\[
 A_h=Dh_{x_\star}H^{-1}\Sigma^{1/2}.
\]
Suppose $A_h$ has row rank $q$; $\Sigma$ itself need not be positive definite. Let
\[
 D_n=D(h\circ\Exp_{a_n})(\bar v_n),
 \qquad
 \widehat V_{h,n}=D_n\widehat V_nD_n^\transpose.
\]
Then
\begin{equation}\label{eq:functional-clt}
 \sqrt{N_n}\{h(\widehat x_n)-h(x_\star)\}
 \rightsquig
 \cN\left(0,A_hA_h^\transpose\right),
\end{equation}
$\widehat V_{h,n}$ is invertible with probability tending to one, and
\begin{equation}\label{eq:functional-pivot}
 N_n\{h(\widehat x_n)-h(x_\star)\}^\transpose
 \widehat V_{h,n}^{-1}
 \{h(\widehat x_n)-h(x_\star)\}
 \rightsquig \cT_q.
\end{equation}
\end{corollary}

For a scalar functional, report
\[
 h(\widehat x_n)\ \pm\
 \sqrt{c_{1,1-\alpha}\widehat V_{h,n}/N_n}.
\]
Thus a loading uses a scalar critical value even for a subspace with a large tangent dimension. The rank condition requires nondegenerate variation in the reported functional: directions annihilated by $Dh_{x_\star}$ need not have positive noise variance. This permits inference for an informative contrast even when the full tangent covariance is singular. Conversely, a zero derivative removes the first order fluctuation and calls for a different limit. For example, a squared loading $h_j([u])=u_j^2$ has derivative $2u_j\delta_j$; a loading at zero is therefore outside the scope of this nondegenerate first order result. Consistent numerical derivatives and a vanishing nonnegative ridge preserve the pivot (\Cref{cor:derivative-approximation,prop:ridge-stability}); \Cref{rem:direct-functional} gives direct normalization of a scalar path.

\subsection{Geometric models}\label{sec:applications}
The construction applies to smooth Riemannian $M$-estimation with i.i.d. observations and stochastic gradient $G(x,Z)=\grad_x\ell(x;Z)$. When differentiation and expectation can be interchanged,
\begin{equation}\label{eq:M-H-Sigma}
\begin{gathered}
 H=\E\{\Hess_x\ell(x_\star;Z)\},\\
 \Sigma=\E\{\grad\ell(x_\star;Z)\ot\grad\ell(x_\star;Z)\}.
\end{gathered}
\end{equation}
We verify Assumptions~\ref{ass:geometry}--\ref{ass:fclt} for the following models under the localization and mean square stability conditions in their respective propositions. Smooth local geometry permits the charts in Assumption~\ref{ass:geometry}. The Hessian and moment calculations below establish Assumptions~\ref{ass:objective} and \ref{ass:oracle}, while i.i.d. true-point gradients with finite second moments give Assumption~\ref{ass:fclt}. For Assumption~\ref{ass:stability}, choose the gain exponent in \eqref{eq:beta-range} and use the stability rate required by each model proposition, or verify it through Proposition~\ref{prop:stability}. Thus localization and stability remain conditions on the recursion. The precise model statements are in \Cref{app:geometric-results}; initialization is deterministic or independent of the observation stream.

\subsubsection{Intrinsic Fr\'echet means}\label{sec:frechet}
For a manifold-valued observation $Z$, the intrinsic mean minimizes
\begin{equation*}
 F(x)=\frac12\E\{\dist(x,Z)^2\}.
\end{equation*}
Away from the cut locus, the stochastic gradient is
\begin{equation}\label{eq:frechet-gradient}
 G(x,Z)=-\Log_x(Z).
\end{equation}
Thus the update moves toward the next observation along its logarithmic direction. At the unique mean $\mu$, the operators in \eqref{eq:M-H-Sigma} become
\begin{equation}\label{eq:frechet-H-Sigma}
\begin{gathered}
 H_\mu=\E\left\{
 \left.\Hess_x\frac12\dist(x,Z)^2\right|_{x=\mu}
 \right\},\\
 \Sigma_\mu=\E\{\Log_\mu(Z)\ot\Log_\mu(Z)\}.
\end{gathered}
\end{equation}
Positive definiteness of $H_\mu$ gives regular local identification. A compact strongly convex support separated from the relevant cut loci, with a square integrable local Hessian/H\"older envelope, is sufficient for smoothness and noise continuity. On a Hadamard manifold, the logarithm is globally defined and finite squared distance gives uniqueness; the local moment and smoothness conditions then verify the remaining model assumptions. \Cref{prop:frechet} states the precise alternatives. These means summarize directional, shape, and matrix-valued observations \citep{BhattacharyaPatrangenaru2005,Pennec2006,MardiaJupp2000}.

\paragraph{Spherical means.}
On the unit sphere $\mathbb S^q$, put $r=\dist(x,z)$ and $u=\Log_x(z)/r$ for $0<r<\pi$. The Hessian of squared distance is
\begin{equation}\label{eq:sphere-hessian}
 \Hess_x\frac12\dist(x,z)^2
 =u\ot u+r\cot(r)(\Id-u\ot u).
\end{equation}
At $r=0$, its continuous extension is $\Id$. Formula \eqref{eq:sphere-hessian} separates the radial eigenvalue, equal to one, from the transverse eigenvalues $r\cot(r)$. Support in a geodesic ball of radius less than $\pi/4$ provides a simple sufficient condition for uniqueness and local strong convexity \citep{Afsari2011}. Averaging \eqref{eq:sphere-hessian} gives $H_\mu$, while logarithmic outer products give $\Sigma_\mu$. \Cref{alg:one-path} obtains the confidence region directly from the iterate path.

\paragraph{SPD matrix means.}
For $\mathbb S_{++}^p$ with affine invariant metric
$\ip{U}{V}_X=\tr(X^{-1}UX^{-1}V)$, the logarithm is
\begin{equation*}
 \Log_X(Z)
 =X^{1/2}\log(X^{-1/2}ZX^{-1/2})X^{1/2}.
\end{equation*}
This is a Hadamard manifold with tangent dimension $p(p+1)/2$. Finite squared affine invariant distance gives a unique mean $M$. Bounded geodesic support implies the local smoothness and noise-continuity bounds. For unbounded support, the local mean-field and noise-continuity conditions in \Cref{prop:frechet}(b) are imposed separately. The covariance factor is $H_M^{-1}\Sigma_MH_M^{-1}$, with both operators defined by \eqref{eq:frechet-H-Sigma}; \eqref{eq:frechet-clt} gives the corresponding limit. Self-normalization avoids differentiating the matrix logarithm for Hessian estimation and transporting gradient outer products to an estimated mean. Such means are used in diffusion tensor and covariance valued analysis \citep{PennecFillardAyache2006,FletcherJoshi2007,Dryden2009}.

\subsubsection{Streaming principal subspaces}\label{sec:pca}
Let $C=\E(XX^\transpose)$ for $X\in\R^p$. Its leading $r$-dimensional eigenspace is a point on $\operatorname{Gr}(r,p)$, with tangent dimension $r(p-r)$ \citep{Edelman1998}. For an orthonormal representative $U$, use
\begin{equation}\label{eq:pca-objective}
 F([U])=-\frac12\E\norm{U^\transpose X}^2
 =-\frac12\tr(U^\transpose CU),
\end{equation}
and stochastic gradient
\begin{equation}\label{eq:pca-gradient}
 G([U],X)=-(\Id-UU^\transpose)XX^\transpose U.
\end{equation}
The operator $C$ is the second moment; it is the covariance when $X$ is centered. Incremental gradient methods include Oja's update and Grassmann subspace tracking \citep{Oja1982,Balzano2010}; averaged Riemannian gradients give an efficient streaming $k$-PCA estimate \citep{Tripuraneni2018}.

Write the spectral decomposition as
\[
 C=U_\star\Lambda_1U_\star^\transpose
   +U_\perp\Lambda_2U_\perp^\transpose,
 \qquad \lambda_r>\lambda_{r+1}.
\]
Here $\lambda_1\ge\cdots\ge\lambda_p$ are the eigenvalues of $C$, and $\Lambda_1,\Lambda_2$ contain the first $r$ and remaining eigenvalues. Represent a tangent perturbation at $[U_\star]$ by $U_\perp K$, with $K\in\R^{(p-r)\times r}$ and the Frobenius inner product. The Hessian acts entrywise:
\begin{equation}\label{eq:pca-H}
 (H_{\mathrm{PCA}}K)_{ij}
 =(\lambda_j-\lambda_{r+i})K_{ij}.
\end{equation}
Hence the eigengap is the smallest Hessian eigenvalue and identifies the leading subspace. With $X_\parallel=U_\star^\transpose X$ and $X_\perp=U_\perp^\transpose X$, the tangent noise covariance is
{
\begin{equation}\label{eq:pca-Sigma}
 \Sigma_{\mathrm{PCA}}=\E\big[
 \vecop(X_\perp X_\parallel^\transpose)
 \vecop(X_\perp X_\parallel^\transpose)^\transpose\big].
\end{equation}
}

Since $\norm{G([U],X)}\le\norm X^2$, the fourth data moment $\E\norm X^4<\infty$ ensures a finite second moment of the stochastic gradient. Under localization and mean square stability, \Cref{prop:pca} gives the root-$N_n$ Gaussian limit for $1/2<\beta<1$. A positive definite tangent covariance yields inference for the full subspace; a nondegenerate derivative contrast yields smooth functional inference. For $r=1$, the parameter is an unoriented direction, making squared loadings natural targets for \Cref{sec:wine-application}.

\begin{table*}[t]
\caption{Nominal 95\% scalar coverage. Signed-H permits negative nonzero estimated Hessians; Oracle uses population asymptotic covariance.}\label{tab:scalar-coverage}
\centering\small
\begin{tabular}{lrcccc}
\toprule
Setting & $n$ & Self-normalized & Plug-in & Signed-H & Oracle \\
\midrule
Multi-frequency, cap 100 & 4000 & 0.928 & 0.846 & 0.870 & 0.818 \\
Multi-frequency, cap 100 & 12000 & 0.948 & 0.900 & 0.900 & 0.858 \\
\addlinespace
Multi-frequency, cap 200 & 4000 & 0.916 & 0.798 & 0.884 & 0.834 \\
Multi-frequency, cap 200 & 12000 & 0.950 & 0.880 & 0.884 & 0.856 \\
\addlinespace
Phase retrieval, mixture & 4000 & 0.934 & 0.944 & -- & 0.960 \\
Phase retrieval, mixture & 12000 & 0.944 & 0.938 & -- & 0.956 \\
\addlinespace
Range localization, mixture & 4000 & 0.918 & 0.918 & -- & 0.944 \\
Range localization, mixture & 12000 & 0.926 & 0.914 & -- & 0.944 \\
\bottomrule
\end{tabular}

\end{table*}

\begin{table*}[t]
\caption{Nominal 95\% full-region coverage and mean squared geodesic error for intrinsic means. Methods share the estimator.}\label{tab:formal-results}
\centering\small
\begin{tabular}{lrrccccc}
\toprule
Setting & $n$ & $d$ & Self-normalized & Short-batch & Population & Plug-in & MSE\\
\midrule
Sphere & 5000 & 2 & 0.922 & 0.646 & 0.940 & 0.936 & $2.92\times10^{-5}$\\
Sphere & 10000 & 2 & 0.926 & 0.658 & -- & -- & $1.44\times10^{-5}$\\
SPD & 5000 & 3 & 0.922 & 0.626 & 0.932 & 0.930 & $5.55\times10^{-5}$\\
SPD & 10000 & 3 & 0.920 & 0.602 & -- & -- & $2.62\times10^{-5}$\\
\bottomrule
\end{tabular}
\end{table*}

\section{Simulations}\label{sec:simulations}
\subsection{Simulation setup}
We compare normalization of one path with plug-in Wald inference in four settings: two multi-frequency circular models, conditional-median phase retrieval on $\mathbb{RP}^2$, and conditional-median range localization on $\mathbb S^2$. The first pair varies the maximum frequency while keeping the population Hessian and score variance fixed; the sensing models require residual-density estimation for the Hessian. Each setting uses $n=4000,12000$ and 500 independent replications, with methods paired on the same averaged estimate. The gain exponent is $0.6$ and $b_n=\lceil n^{1/3}\rceil$. We report nominal 95\% coverage and mean absolute error for the scalar functional. Failures count as noncoverage. Plug-in uses analytic sample Hessians for the circular models and bandwidth multiplier one for sensing; Signed-H accepts any nonzero scalar Hessian. Oracle uses population asymptotic covariance. Appendix~\ref{app:scalar-designs} gives the data-generating models and inference rules.

In the circular models, $\theta_\star=0.35$ and $\Phi=M\theta_\star+\epsilon$ modulo $2\pi$. Frequency $M$ equals one with probability $0.7$ and otherwise is $\min\{m_{\max},\lfloor20U^{-2/3}\rfloor\}$ for uniform $U$, with caps 100 and 200. The conditional phase density is $\{1+(1.2/M)\cos\epsilon+0.4\cos(2\epsilon)\}/(2\pi)$. The loss $\{1-\cos(M\theta-\Phi)\}/M$ has bounded score and population operators $H=0.6$, $\Sigma=0.4$; its observation Hessian grows with frequency. We infer $\sin\theta_\star$, use gain constant $0.8$, and project updates onto $[-0.4,1]$.

The sensing observations are $Y=f(x_\star,A)+\sigma(A)(Z-m_Z)$, where $Z\sim0.8\cN(0,1)+0.2\cN(0.5,0.15^2)$ and $m_Z$ is its median. Half absolute residual loss targets the conditional median. Phase retrieval uses $f(x,A)=(A^\transpose x)^2$, a uniform spherical design, and target $x_1^2$ on $\mathbb{RP}^2$; range localization uses $f(x,A)=\norm{x-A}$, six axis-aligned sensors, and target $x_1$ on $\mathbb S^2$. These exercise nonlinear coordinates, whereas the circular chart is locally Euclidean. Appendix~\ref{app:scalar-designs} specifies scales, initializations, gain constants, and the density-based plug-in rule.

We also study full-region inference for intrinsic means with readily estimated Hessians. Each setting uses 500 independent paths. The sphere model is rotationally symmetric with concentration eight, truncated to a geodesic ball of radius $0.6$ about its mean. The positive definite model is $Z=M^{1/2}\exp(A)M^{1/2}$, where $M$ has condition number five and the orthonormal symmetric coordinates of $A$ are independent $N(0,0.3^2)$. Both models use exact exponential and logarithm maps. Gains are $2k^{-0.65}$ on the sphere and $1.5k^{-0.6}$ on the SPD matrix cone; burn-in is $\lceil n^{1/3}\rceil$.

The settings were selected after exploratory comparisons. The cap-100 design was then evaluated with a separately fixed independent seed set; the other primary settings reuse their fixed simulation runs. Additional comparisons include single-frequency controls and settings with lower self-normalized coverage.

\begin{table}[tb]
\caption{Largest squared white wine loadings and nominal marginal 95\% intervals.}\label{tab:wine-loadings}
\centering
\setlength{\tabcolsep}{3.5pt}
\begin{tabular}{lrr}
\toprule
Variable & Squared loading & 95\% interval\\
\midrule
Density & 0.217 & $[0.201,0.233]$\\
Alcohol & 0.199 & $[0.191,0.207]$\\
Total sulfur dioxide & 0.177 & $[0.156,0.197]$\\
Residual sugar & 0.165 & $[0.145,0.185]$\\
Free sulfur dioxide & 0.102 & $[0.071,0.132]$\\
Chlorides & 0.056 & $[0.030,0.082]$\\
\bottomrule
\end{tabular}

\end{table}

\subsection{Results}

\Cref{tab:scalar-coverage} shows the largest coverage gain when observation Hessians fluctuate strongly. Accepting negative nonzero Hessian estimates improves short-stream plug-in coverage, but the longer-stream advantage remains. Population-covariance Wald intervals also undercover, so covariance estimation alone does not explain the difference. Sensing gives comparable coverage without residual-density estimation; estimation errors decrease with stream length (\Cref{fig:scalar-error}).

In the intrinsic-mean comparisons, estimation error decreases while self-normalized coverage remains below nominal (\Cref{tab:formal-results}). Plug-in performs better at $n=5000$ and in most matched scalar controls (\Cref{tab:additional-scalar}). Short-batch uses $\lfloor\sqrt{N_n}\rfloor$ consecutive batches and serves as a failure diagnostic, not an adapted batch-means procedure. Population Wald uses independently approximated population operators; plug-in uses observation-based operators at the fitted estimate (Appendix~\ref{app:scalar-designs}). These comparisons distinguish the gain under strongly varying observation Hessians from settings where curvature is readily estimated. Scalar projection reduces dimension but does not by itself ensure accurate coverage.

\label{sec:wine-application}
The data illustration analyzes 4898 white wine samples with 11 physicochemical measurements from the UCI Wine Quality data \citep{CortezUCI2009,Cortez2009}. A preprocessing block of 898 rows fixes centers and scales; the remaining 4000 rows form a randomly ordered PCA stream targeting the leading standardized second moment eigenspace conditional on preprocessing. We use $\eta_k=0.15k^{-0.6}$, burn-in 16, and a fixed initial direction. The estimate lies $0.0127$ radians from the batch eigenspace and explains $28.3\%$ of the standardized second moment; the sample eigengap is $1.441$.

We infer squared loadings $h_j([u])=u_j^2$ using \Cref{cor:functionals}, projecting the ten-dimensional normalizer onto each squared loading. A common anchor graph chart defines the path average, centered sums, and functional derivatives (Appendix~\ref{app:wine-details}). Density, alcohol, total sulfur dioxide, and residual sugar have the largest squared loadings (\Cref{tab:wine-loadings}); intervals are intersected with $[0,1]$. Under an i.i.d. sampling model these intervals are asymptotically marginal for fixed nondegenerate loadings, conditional on preprocessing. They describe physicochemical variation, excluding sensory quality; ranking the loadings does not confer simultaneous or selective coverage. Appendix~\ref{app:wine-details} reports derivative and order sensitivity.

\section{Discussion}\label{sec:conclusion}
One Riemannian gradient path yields confidence regions in one pass without sandwich covariance estimation. The joint endpoint and $L^2$ expansion justifies using Brownian critical values under stability and martingale conditions, avoiding highly variable Hessian estimates and residual-density estimation.

Finite-sample undercoverage remains in some settings. Quantitative martingale approximations, geometric remainder bounds, and control of small normalizer eigenvalues could guide gain selection and coverage corrections. The present theory also assumes fixed dimension, local stability, and nondegenerate parameters or functionals at deterministic sample sizes. For increasing dimension or weak identification, projection and regularization are promising directions.

\clearpage
\section*{AI use statement}
Generative AI tools were used for language editing, literature searches, code development, and checking the authors’ mathematical arguments. The authors verified all AI-assisted material incorporated into the manuscript and code, and take full responsibility for the work.

\clearpage
\section*{Supplementary Materials}
\appendix
\numberwithin{equation}{section}

\section{Additional Experiments}\label{app:simulation-details}
\subsection{Brownian critical values}\label{sec:calibration}
\Cref{sec:calibration-coordinates} describes simulation of dimension-dependent Brownian critical values. \Cref{tab:brownian-calibration} lists the critical values for the main and supplementary experiments.
Every dimension uses 20000 independent internal draws. Dimensions $1,2,3,5,6$ use 512 Karhunen--Lo\`eve modes; dimensions $10,15,19$ use 8192 modes. The scalar critical value rounds to $46.14$; calculations use the unrounded critical value. These internal draws are separate from the 500 outer replications per setting.

\begin{table}[ht]
\caption{Brownian 0.95 critical values.}\label{tab:brownian-calibration}
\centering\begin{tabular}{rrrrrrrrr}
\toprule
$d$ & 1 & 2 & 3 & 5 & 6 & 10 & 15 & 19\\
\midrule
$c_{d,0.95}$ & 46.14 & 100.79 & 172.91 & 366.61 & 477.04 & 1017.28 & 1972.35 & 2917.21\\
\bottomrule
\end{tabular}

\end{table}

\subsection{Scalar models and inference rules}\label{app:scalar-designs}

The four primary settings separate two obstacles to covariance estimation: a highly variable observation Hessian and a population Hessian that depends on an unknown density. Each uses 500 independent paths at each of $n=4000,12000$. Gains are $\eta_k=\eta_0k^{-0.6}$, with $b_n=\lceil n^{1/3}\rceil$ and $N_n=n-b_n$. The retained sample sizes are 3984 and 11977. All methods use the same fixed-chart averaged estimate, the same $N_n$ scaling, and no ridge. Self-normalization uses \Cref{cor:functionals}; the derivative is analytic in the circular models and approximated by centered coordinate differences with step $10^{-3}n^{-1/4}$ in the sensing models. Estimation error is $\abs{h(\widehat x_n)-h(x_\star)}$.

\paragraph{Multi-frequency circular estimation.}
Observe an integer frequency $M$ and an angle $\Phi=M\theta_\star+\epsilon$ modulo $2\pi$, with $\theta_\star=0.35$. Set $M=1$ with probability $0.7$; otherwise take $M=\min\{m_{\max},\lfloor20U^{-2/3}\rfloor\}$ for an independent uniform $U$. The two frequency caps are $m_{\max}=100,200$. Conditional on $M$, the phase error has density
\[
 p(\epsilon\mid M)=\frac{1+(1.2/M)\cos\epsilon+0.4\cos(2\epsilon)}{2\pi},
 \qquad -\pi<\epsilon\le\pi.
\]
For loss $\ell(\theta;M,\Phi)=\{1-\cos(M\theta-\Phi)\}/M$, the score is $\sin(M\theta-\Phi)$ and its derivative is $M\cos(M\theta-\Phi)$. Thus the score is bounded by one, whereas the observation Hessian scales with frequency. Direct integration gives $H=0.6$ and $\Sigma=0.4$ for either cap. Increasing the cap changes observation-Hessian variability without changing these population operators. In particular, at the true parameter the second moment of the observation Hessian is $0.6\E(M^2)$, while its mean remains $0.6$.

The reported functional is $h(\theta)=\sin\theta$. Starting at zero with $\eta_0=0.8$, we project each angular update onto $[-0.4,1]$ and average in the anchor's angular chart, denoting the estimate by $\widehat\theta_n$. The positive-Hessian plug-in uses all $n$ observations to estimate
\[
 \widehat H=n^{-1}\sum_i M_i\cos(M_i\widehat\theta_n-\Phi_i),\qquad
 \widehat\Sigma=n^{-1}\sum_i(g_i-\bar g)^2,
 \quad g_i=\sin(M_i\widehat\theta_n-\Phi_i),
\]
where $\bar g=n^{-1}\sum_i g_i$. The scalar sandwich is $\cos^2(\widehat\theta_n)\widehat\Sigma/\widehat H^2$, provided $\widehat H>10^{-9}$. It requires no bandwidth. Signed-H uses the same sandwich whenever $\abs{\widehat H}>10^{-9}$, since scalar variance does not require a positive estimated Hessian; both rules appear in \Cref{tab:scalar-coverage}. The oracle uses the population asymptotic variance $\cos^2(0.35)\,0.4/0.6^2$, not the finite-sample distribution of the estimator.

The caps keep $H$ and $\Sigma$ fixed, but also change the mean field away from the true parameter and the score-continuity constants. Thus they vary observation Hessians within regular models without holding the entire finite-stream distribution fixed. Angular coordinates are locally Euclidean here; the sensing examples exercise nonlinear manifold coordinates.

Identification and stability can be checked directly. For $\delta=\theta-\theta_\star$,
\[
 F(\theta)-F(\theta_\star)=\E\left[\frac{0.6}{M^2}\{1-\cos(M\delta)\}\right].
\]
The positive mass at $M=1$ identifies $\theta_\star$ uniquely on the circle. On the projection interval, $\abs\delta\le0.75$ and
$F''(\theta)=0.6\E\cos(M\delta)\ge0.6\{0.7\cos(0.75)-0.3\}>0$.
Nonexpansiveness of projection and the bounded score give the quadratic drift
\[
 \E\{(\theta_{k+1}-\theta_\star)^2\mid\F_k\}
 \le(1-2\mu_0\eta_{k+1})(\theta_k-\theta_\star)^2+\eta_{k+1}^2,
 \quad \mu_0=0.6\{0.7\cos(0.75)-0.3\}.
\]
This yields $\E(\theta_k-\theta_\star)^2=O(\eta_k)$ and almost-sure convergence by the decreasing-step supermartingale argument. Since the true parameter is interior and the score is bounded, projection is eventually inactive. In a smaller angular neighborhood, the reset auxiliary recursion has strongly monotone drift inside the neighborhood and linear drift outside, with bounded noise. The same drift bound and the burn-in moment bound verify \Cref{prop:stability}. Finite caps give bounded derivatives and $\rho=\nu=1$; the i.i.d. bounded true-point scores verify the martingale conditions. Hence this experiment has a regular, identifiable model with strongly varying observation Hessians rather than a singular population Hessian.

\paragraph{Conditional-median sensing.}
Both sensing models observe $Y=f(x_\star,A)+\sigma(A)(Z-m_Z)$ and minimize half absolute residual loss. Let $\Phi_{\mathrm N}$ and $\phi_{\mathrm N}$ denote the standard normal distribution function and density. The noise variable and its median $m_Z$ satisfy
\[
 Z\sim 0.8\cN(0,1)+0.2\cN(0.5,0.15^2),\qquad
 0.8\Phi_{\mathrm N}(m_Z)+0.2\Phi_{\mathrm N}\{(m_Z-0.5)/0.15\}=0.5,
\]
so $m_Z\simeq0.2749$ and the conditional median is $f(x_\star,A)$. The asymmetric mixture introduces density variation near the median without shifting the conditional median.

For known-energy phase retrieval, a unit representative $x\in\mathbb S^2$ denotes a point in $\mathbb{RP}^2$. We take $A$ uniform on the sphere of radius $\sqrt3$,
\[
 f(x,A)=(A^\transpose x)^2,\qquad
 x_\star=(0.4,-0.3,1)^\transpose/\sqrt{1.25},\qquad
 \sigma(A)=0.2(1+0.3A_1/\sqrt3).
\]
The sign-invariant functional is $h(x)=x_1^2$, with true value $0.128$. Initialization is $(0.2,-0.1,1)^\transpose/\sqrt{1.05}$, $\eta_0=0.2$, and representatives are aligned before taking projective logarithms. Measurements follow the additive model above, as for background-corrected intensities. Population operators for the oracle are integrated over the spherical design; doubling Gauss--Legendre order from 32 to 64 changes the scalar variance by less than $10^{-9}$ relatively.

For range localization, $x\in\mathbb S^2$ is a position on a known spherical surface. A sensor $A$ is chosen uniformly from $(2e_1,2e_2,2e_3,-2e_1,-2e_2,-2e_3)$, where $e_j$ are the standard coordinate vectors, with prediction $f(x,A)=\norm{x-A}$. We set $x_\star=\Exp_{e_3}(0.2,-0.15,0)$ and take the six noise scales to be $0.08$ times six equally spaced values from $0.7$ to $1.3$, in that sensor order. The reported coordinate is $h(x)=x_1$, approximately $0.1979$. Initialization is $e_3$ and $\eta_0=0.8$. Opposite-sensor squared ranges identify $x_j$ through $\{\norm{x+2e_j}^2-\norm{x-2e_j}^2\}/8$; all sensor distances are at least one. The oracle is computed by an exact six-sensor sum.

Both models use exponential-map updates. Write $r_i=Y_i-f(\widehat x_n,A_i)$, $s_i=\ind\{r_i<0\}-0.5$, and let $D_i$ be the tangent gradient of $f$ at $\widehat x_n$. The plug-in estimates
\begin{align*}
 \widehat H&=n^{-1}\sum_i\left\{\frac{\phi_{\mathrm N}(r_i/t_n)}{t_n}D_iD_i^\transpose+s_i\Hess_x f(\widehat x_n,A_i)\right\},\\
 \widehat\Sigma&=n^{-1}\sum_i(s_iD_i-\overline{sD})(s_iD_i-\overline{sD})^\transpose,\qquad
 \overline{sD}=n^{-1}\sum_i s_iD_i,\\
 t_n&=1.4826\operatorname{median}_i\abs{r_i-\operatorname{median}_j r_j}\;n^{-1/5}.
\end{align*}
The bandwidth multiplier is one for every sensing setting. The geometric correction in $\widehat H$ is retained, and inverse action on the functional derivative gives the scalar sandwich variance. Nonpositive Hessians or variances count as failed intervals, without substitution of a population value. These are all-observation comparisons at the common fitted estimate; neither plug-in nor oracle changes that estimate.

To connect these losses to the theory, let $D(A)=\grad_x f(x_\star,A)$ and $p_\epsilon(0\mid A)$ be the conditional density at zero of $\epsilon=Y-f(x_\star,A)$. Differentiating the conditional mean score gives
\[
 H=\E\{p_\epsilon(0\mid A)D(A)\ot D(A)\},\qquad
 \Sigma=\tfrac14\E\{D(A)\ot D(A)\}.
\]
Thus the population Hessian contains a density term even though a single observation's indicator score is locally constant off its threshold. Smooth predictors, positive smooth error densities and the spanning designs make $H$ positive definite and give a quadratic mean-field remainder. An indicator changes only on a residual interval of length proportional to the parameter displacement, giving squared $L^2$ score increments of that order. These verify $\rho=1$ and $\nu=1/2$, while bounded scores verify the true-point martingale conditions. \Cref{cor:functionals} then applies under the recursion localization and uniform mean square stability conditions in Assumptions~\ref{ass:geometry} and \ref{ass:stability}, with $\beta=0.6$.

\paragraph{Full-region comparison rules.}
The intrinsic-mean comparisons in \Cref{tab:formal-results} use the same fitted estimates and retained sample size $N=N_n$. The short-batch diagnostic takes $B=\max\{4,\lfloor\sqrt N\rfloor\}$ consecutive batches of length $\ell=\lfloor N/B\rfloor$ from the retained coordinate path. Writing $\bar v_j^{\,B}$ for batch means and $\bar v^{\,B}$ for their average,
\[
 \widehat\Omega_B=\frac{\ell}{B-1}\sum_{j=1}^{B}
 (\bar v_j^{\,B}-\bar v^{\,B})(\bar v_j^{\,B}-\bar v^{\,B})^\transpose.
\]
The Wald statistic uses the full retained average, scale $N$, and $\chi_d^2$ critical value. This short-batch schedule is a failure diagnostic, not the adapted batch-means procedure of \citet{Chen2020}.

Population Wald regions use independently approximated population Hessians and covariances at the true parameter, mapped into the anchor chart. Plug-in regions estimate these operators at the fitted point using observations: all observations for the sphere, and an 80-observation Hessian approximation together with gradient covariance from all observations for positive definite matrices. These regions also use $\chi_d^2$ critical values.

\subsection{Estimation error and covariance comparisons}
\begin{figure}[ht]
\centering\includegraphics[width=\textwidth]{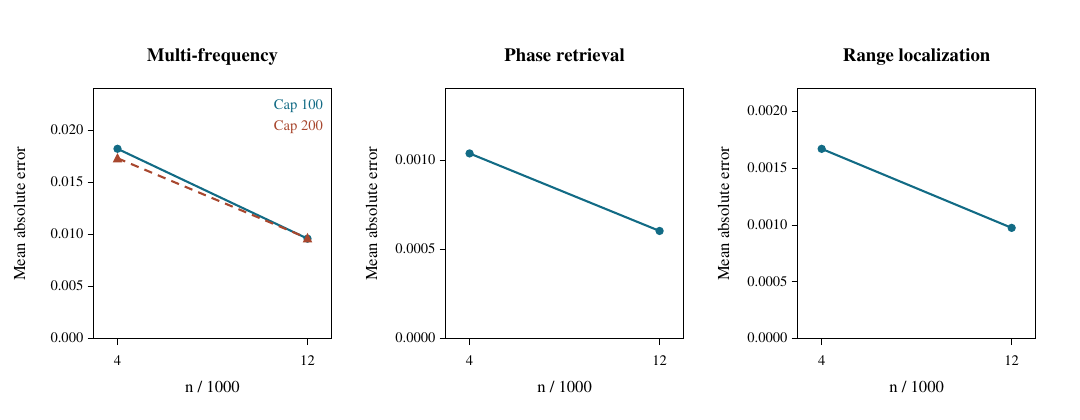}
\caption{Mean absolute error of the common scalar estimator from 500 replications per setting. Each panel has its own vertical scale.}\label{fig:scalar-error}
\end{figure}
The multi-frequency mean errors decrease from $0.0182$ to $0.00961$ at cap 100 and from $0.0173$ to $0.00966$ at cap 200. Phase-retrieval errors decrease from $0.00104$ to $0.00060$, and range-localization errors from $0.00167$ to $0.00098$. These are shared point-estimation results, not differences between the inference methods.

The circular plug-in fails its positive-Hessian check in 12 and 0 replications at cap 100 and in 43 and 2 replications at cap 200, in increasing order of $n$. All 500 replications remain in every coverage denominator. No self-normalized, signed-Hessian or oracle calculation fails, and neither method fails in the selected sensing experiments. Allowing negative nonzero Hessians improves short-stream plug-in coverage, but the longer-stream coverage gap remains (\Cref{tab:scalar-coverage}).

\paragraph{Single-frequency and other sensing comparisons.}
Setting $M=1$ throughout gives a single-frequency control with the same conditional phase density, reported functional, gain, initialization and projection as the multi-frequency models. The two sensing models are also evaluated with $Z\sim\cN(0,1)$ and $m_Z=0$, leaving their other settings unchanged. \Cref{tab:additional-scalar} reports these comparisons together with circular intrinsic means. The single-frequency control at $n=12000$ favors plug-in, while phase retrieval remains close under both error distributions. Consequently the gain in the main frequency comparison occurs with strongly varying observation Hessians, not across all circular or density-dependent models.

For the circular intrinsic means, $Y=0.35+T$ modulo $2\pi$ with
\[
 p_m(t)=\{1+0.6\cos t+0.1\cos(mt)+0.05(\sin t+2\sin2t)\}/(2\pi),\qquad m=2,8.
\]
Half squared geodesic loss has target mean $0.35$, Hessian $H=0.5$, and score variance $\pi^2/3-1.2+0.2/m^2$. The reported mean functional is $\sin(0.35)$; gain, initialization, projection and burn-in equal those in the multi-frequency models. The density at the fitted antipode is estimated by a wrapped Gaussian kernel with bandwidth $n^{-1/5}$, and $1-2\pi\widehat p(\widehat\theta_n+\pi)$ estimates the Hessian. This is a different density-dependent mechanism from the analytic observation Hessian in the multi-frequency experiment.
\begin{table}[ht]
\centering\small
\caption{Additional scalar coverage comparisons, each with 500 replications. All plug-in bandwidth multipliers are one.}\label{tab:additional-scalar}
\begin{tabular}{lrccc}
\toprule
Setting & $n$ & Self-normalized & Plug-in & Oracle \\
\midrule
Single frequency & 4000 & 0.938 & 0.940 & 0.942 \\
Single frequency & 12000 & 0.926 & 0.956 & 0.956 \\
\addlinespace
Intrinsic circle, $m=2$ & 4000 & 0.894 & 0.946 & 0.942 \\
Intrinsic circle, $m=2$ & 12000 & 0.924 & 0.952 & 0.954 \\
\addlinespace
Intrinsic circle, $m=8$ & 4000 & 0.926 & 0.934 & 0.956 \\
Intrinsic circle, $m=8$ & 12000 & 0.920 & 0.938 & 0.970 \\
\addlinespace
Phase retrieval, normal & 4000 & 0.934 & 0.946 & 0.944 \\
Phase retrieval, normal & 12000 & 0.934 & 0.944 & 0.944 \\
\addlinespace
Range localization, normal & 4000 & 0.934 & 0.948 & 0.938 \\
Range localization, normal & 12000 & 0.932 & 0.956 & 0.950 \\
\bottomrule
\end{tabular}

\end{table}

\subsection{Details of the white wine analysis}\label{app:wine-details}
\begin{figure}[ht]
\centering\includegraphics[width=.75\textwidth]{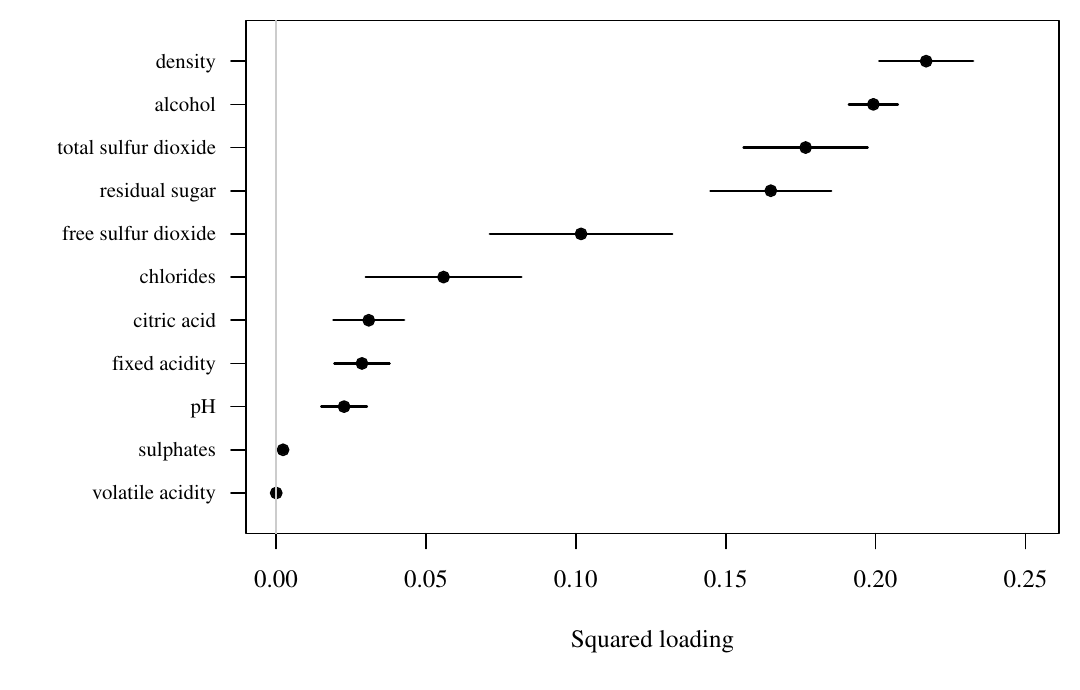}
\caption{Squared white wine loadings with nominal marginal 95\% intervals.}\label{fig:wine-loadings}
\end{figure}

Each row is a sampling unit. The 11 physicochemical variables have no reported missing values; repeated measurement rows are retained. The sensory quality score is excluded. Standardization estimated from the preprocessing block remains fixed for the streaming analysis.
Figure~\ref{fig:wine-loadings} displays the squared-loading intervals. The anchor graph chart uses $z_k=B^\transpose u_k/(a^\transpose u_k)$ and inverse $u_a(z)=(a+Bz)/(1+\norm z^2)^{1/2}$, where $B$ is an orthonormal tangent basis. These coordinates are invariant to changing the sign of $u_k$; choose representatives with $a^\transpose u_k>0$. Their average and centered sums define the estimate and normalizer, and the derivative is evaluated in the same coordinates. For the graph coordinates used in Section~\ref{sec:wine-application}, with $c=a+Bz$, the derivative is
\[
 \nabla_z h_j\{u_a(z)\}=\frac{2c_j}{1+\norm z^2}B^\transpose e_j
 -\frac{2c_j^2}{(1+\norm z^2)^2}z.
\]
The reported calculation uses centered differences with fixed step $10^{-5}$; steps $10^{-4},10^{-5},10^{-6}$ give stable interval widths. This is a numerical precision check, distinct from the shrinking-step asymptotics of \Cref{cor:derivative-approximation}. Across ten additional random orders, the angle to the batch eigenspace has median $0.0157$ and maximum $0.0425$ radians; source-file order gives $0.0596$ radians.

\section{Additional Theory}\label{app:theory-supplement}
This section collects sufficient conditions, functional inference, structural comparisons, and geometric specializations of \Cref{thm:integrated-fclt,thm:selfnorm}. Their proofs are collected in \Cref{app:proofs}.
\subsection{Scope and verification of the assumptions}\label{sec:assumption-discussion}

A common normal neighborhood admits a local logarithm chart. Such localizations are standard in stochastic approximation \citep{KushnerYin2003} and Riemannian gradient theory \citep{Bonnabel2013,Tripuraneni2018}, and follow from almost sure convergence or projection onto a strongly convex set. Hadamard manifolds have globally single-valued logarithms; local boundedness controls Taylor remainders. For intrinsic means, support separated from the cut locus is a convenient sufficient condition, while specialized theory permits mass near the cut locus \citep{HotzLeWood2024}.

Positive definiteness of $H$ is the usual regular identification condition for a root-$n$ limit of the parameter vector. If $H$ is singular, the convergence rate and limit law may change; smeariness of intrinsic means is one example \citep{EltznerHuckemann2019}. Assumption \ref{ass:objective} states only the Taylor remainder used by the proof. The transported H\"older condition \eqref{eq:hessian-holder} implies this remainder bound. The condition $\beta(1+\rho)>1$ controls the accumulated mean field remainder. The inequality $\beta>1/2$ controls the accumulated retraction and curvature terms, while $\beta<1$ controls the Abel boundary for a generic scalar gain. The boundary choice $\beta=1$ generally has a different partial-sum limit and is not covered by the Brownian critical values used here.

Under localization and uniform stability in mean square, the nonlinear remainder bounds use only the conditional second moment of the stochastic gradient. The square integrable Hessian envelope in \eqref{eq:stochastic-hessian-envelope} gives \eqref{eq:noise-continuity} with $\nu=1$ in an i.i.d. smooth loss model. Conditions \eqref{eq:predictable-covariance}--\eqref{eq:lindeberg} are standard martingale functional central limit conditions \citep{HallHeyde1980}. For i.i.d. gradients at $x_\star$, a finite second moment implies them.

The uniform triangular-array rate in \eqref{eq:l2-rate} is the usual local mean square rate for a stable decreasing-step stochastic approximation, imposed on the auxiliary process used by the proof. Proposition~\ref{prop:stability} derives it from a one-step Lyapunov drift and a uniform burn-in initial bound. In a concrete model, the drift is typically verified from local strong monotonicity of the mean gradient, a second moment bound, and either a projected recursion or an established convergence theorem. The sufficient condition uses the increment-truncated map itself to bound the squared update. It therefore requires neither a fourth gradient moment nor a squared bound on the Taylor remainder. The initial moment bound is not implied by consistency of the anchor.

{
These conditions follow the established averaging framework. In Euclidean random scaling, \citet[Assumption~1]{Lee2022} impose a positive definite Hessian, a local H\"older expansion of the mean gradient, polynomial gains with exponent in $(1/2,1)$, and martingale noise with covariance and tail conditions. These correspond to Assumptions \ref{ass:objective}--\ref{ass:fclt}. Their Lyapunov condition implies stability; here retained localization and a uniform mean square rate state that input directly. Their uniform path argument additionally assumes a higher noise moment depending on the gain exponent. Our joint $L^2$ and endpoint analysis controls nonlinear remainders through conditional second moments, with the martingale covariance and Lindeberg requirements in \Cref{ass:fclt}.

For Riemannian averaging, \citet[Assumptions~1--6]{Tripuraneni2018} require iterates in a compact common retraction neighborhood, a positive definite Hessian with local Lipschitz regularity, conditionally unbiased gradients, continuity in mean square, and a bounded conditional fourth gradient moment. Their Assumption~1 gives $\E\norm{\Delta_k}^2=O(\eta_k)$, the same rate used in \eqref{eq:l2-rate}. The choices $\rho=1$ and $\nu=1$ recover the usual Lipschitz conditions and gain range. Our formulation assumes retained localization and uniform stability and uses the integrated path expansion to connect the random anchor chart to the Brownian normalizer. Thus the inferential assumptions retain the standard local identification, stability, and martingale structure of averaging theory, with conditions tailored to the joint endpoint and path limit.
\par}

A uniform functional limit is still available under an additional maximal condition. For example, if
\begin{equation*}
 \frac{1}{\sqrt{N_n}}
 \max_{b_n\le k\le n}\eta_k^{-1}\norm{\Delta_k}
 \xrightarrow{p}0,
\end{equation*}
then the proof of \Cref{thm:integrated-fclt} yields convergence in the Skorokhod topology and, because the limit is continuous, in the uniform topology.

\subsection{Derivative approximation and regularization}

The rank condition in \Cref{cor:functionals} is the relevant nondegeneracy requirement for a reported contrast. It can hold even when the full tangent covariance is singular.

\begin{corollary}[Consistent derivative approximation]\label{cor:derivative-approximation}
Under the conditions of \Cref{cor:functionals}, let $\widehat D_n:T_{a_n}\M\to\R^q$ be a random linear map such that, after the natural tangent identifications,
\[
 \norm{\widehat D_n-D_n}_{\op}\xrightarrow{p}0.
\]
Then \eqref{eq:functional-pivot} remains valid with $\widehat V_{h,n}$ replaced by
\[
 \widehat V^{\rm app}_{h,n}
 =\widehat D_n\widehat V_n\widehat D_n^\transpose.
\]
In fixed dimension, centered coordinate differences with deterministic steps $s_n\to0$ satisfy the operator-norm consistency condition with error $O_p(s_n^2)$ when the relevant pullbacks are uniformly $C^3$ on a common neighborhood containing every perturbed evaluation point. No rate linking $s_n$ to $N_n$ is required.
\end{corollary}

\begin{proposition}[Stability under a vanishing ridge]\label{prop:ridge-stability}
Suppose the joint endpoint and normalizer convergence in \Cref{thm:selfnorm} holds, and let all ridges below be nonnegative and possibly data-dependent.
\begin{enumerate}[label=(\alph*),leftmargin=2.1em]
\item If $\Sigma$ is positive definite and $\lambda_n\xrightarrow{p}0$, replacing $\widehat V_n$ by $\widehat V_n+\lambda_n\Id$ leaves the full statistic's limit $\cT_d$ unchanged.
\item Under \Cref{cor:functionals}, if $\lambda_{h,n}\xrightarrow{p}0$, replacing the projected normalizer by $\widehat V_{h,n}+\lambda_{h,n}\Id_q$, or replacing $\widehat V^{\rm app}_{h,n}$ analogously under \Cref{cor:derivative-approximation}, leaves the limit $\cT_q$ unchanged.
\end{enumerate}

\end{proposition}

\begin{remark}[Direct normalization of a scalar path]\label{rem:direct-functional}
For a $C^2$ scalar $h$, the arithmetic mean and bridge normalizer of $h(x_{b_n+1}),\ldots,h(x_n)$ give a direct scalar statistic. On the agreement event in \Cref{lem:localization-transfer}, the Taylor remainder is $O(\norm{\Delta_{k,n}^{\rm loc}}^2)$. Event splitting, \eqref{eq:l2-rate}, and $\beta>1/2$ make its accumulated contribution $o_p(1)$ after $N_n^{-1/2}$ scaling. A nondegenerate derivative contrast therefore gives the limit $\cT_1$, equivalent to first order to the construction using derivative projection in \Cref{cor:functionals}.
\end{remark}

\subsection{Structural properties and comparison with Wald inference}\label{sec:formal-comparison}

\begin{proposition}[Coordinate equivariance and Euclidean reduction]\label{prop:coordinate-reduction}
Under Assumptions \ref{ass:geometry}--\ref{ass:fclt}, the following statements hold.
\begin{enumerate}[label=(\alph*),leftmargin=2.1em]
\item The finite sample statistic $T_n(x)$ is unchanged by an orthonormal change of basis in $T_{a_n}\M$.
\item Let $\mathcal A\subset U_0$ be a neighborhood of $x_\star$. For $a\in\mathcal A$, let $\chi_a$ be a $C^2$ coordinate chart on a common open set containing $\overline U_0$, with values in $T_a\M$ and $\chi_a(a)=0$. Suppose $(a,x)\mapsto\chi_a(x)$ is jointly $C^2$, with derivatives through second order uniformly bounded for $a\in\mathcal A$ and $x\in\overline U_0$, and that $D\chi_a|_{x_\star}$ and its inverse are uniformly bounded. If $\Log_a(x)$ is replaced throughout \Cref{alg:one-path} by $\chi_a(x)$ and the average is mapped back by $\chi_a^{-1}$, the alternative-chart average still satisfies \eqref{eq:intrinsic-clt}. If $\Sigma$ is positive definite, the full statistic has limit $\cT_d$. More generally, for $h$ under \Cref{cor:functionals}, derivative projection using $D(h\circ\chi_{a_n}^{-1})$ at the alternative-coordinate average has limit $\cT_q$ whenever $A_h$ has row rank $q$, even if $\Sigma$ is singular. This functional conclusion also holds for an operator-norm-consistent approximation to that derivative.
\item If $\M=\R^d$, $\mathcal R_x(v)=x+v$, and $\chi_a(x)=x-a$, after relabeling the retained segment, \Cref{alg:one-path} is exactly the random scaling construction for Polyak--Ruppert averaged stochastic gradient descent in \citet{Lee2022}.
\end{enumerate}
\end{proposition}

Part (b) covers inverse retractions that form a uniformly $C^2$ local chart on the retained localization neighborhood. Smooth coordinate choices give the same pivotal first order limit.

For a formal comparison with plug-in covariance estimation, define $\Omega=H^{-1}\Sigma H^{-1}$ and let $\widehat\Omega_n$ be an operator on $T_{a_n}\M$ such that
\begin{equation}\label{eq:Omega-consistency}
 \mathsf P_{a_n\to x_\star}\widehat\Omega_n\mathsf P_{x_\star\to a_n}
 \xrightarrow{p}\Omega.
\end{equation}
Let $\chi^2_{d,1-\alpha}$ be the $(1-\alpha)$ quantile of $\chi_d^2$. The corresponding local Wald region is
\begin{equation}\label{eq:wald-region}
 \cC^{\rm W}_{n,1-\alpha}
 =\left\{x:
 N_n\ip{\bar v_n-\Log_{a_n}(x)}
 {\widehat\Omega_n^{-1}\{\bar v_n-\Log_{a_n}(x)\}}_{a_n}
 \le \chi^2_{d,1-\alpha}
 \right\}.
\end{equation}
The Wald region \eqref{eq:wald-region} and the self-normalized region are understood to be restricted to their common chart domain. Let $\omega_d=\pi^{d/2}/\Gamma(1+d/2)$ and let $\operatorname{vol}_{a_n}$ denote Lebesgue volume in an orthonormal coordinate system of $T_{a_n}\M$. Because the region radii are $O_p(N_n^{-1/2})$ and the true parameter is an interior point of that domain with probability tending to one, this restriction is asymptotically inactive.

\begin{proposition}[Relation to Wald inference at first order]\label{prop:wald-comparison}
Suppose the assumptions of \Cref{thm:selfnorm} and \eqref{eq:Omega-consistency} hold, $\Sigma$ is positive definite, and $\widehat\Omega_n$ is self-adjoint and positive definite with probability tending to one. Then the Wald statistic converges to $\chi_d^2$, while the self-normalized statistic converges to $\cT_d$. Moreover, with $L=H^{-1}\Sigma^{1/2}$,
\begin{align}
 N_n^{d/2}\operatorname{vol}_{a_n}
 \{\Log_{a_n}(\cC^{\rm W}_{n,1-\alpha})\}
 &\xrightarrow{p}
 \omega_d(\chi^2_{d,1-\alpha})^{d/2}\abs{\det L},
 \label{eq:wald-volume}\\
 N_n^{d/2}\operatorname{vol}_{a_n}
 \{\Log_{a_n}(\cC_{n,1-\alpha})\}
 &\rightsquig
 \omega_dc_{d,1-\alpha}^{d/2}\abs{\det L}\det(Q_d)^{1/2}.
 \label{eq:sn-volume}
\end{align}
Both procedures share the first order covariance factor $\abs{\det L}$; the self-normalized volume also contains the random bridge factor $\det(Q_d)^{1/2}$.
\end{proposition}

The volume limits separate the shared covariance factor from the random Brownian bridge shape. Comparing region efficiency would involve choosing a loss for random confidence sets and common higher-order asymptotic conditions.

Wald inference estimates $\Omega$ and uses chi-square critical values. Self-normalization uses the path's Brownian bridge shape and its pivotal critical values. Other Euclidean studentization procedures use batch means, perturbed trajectories, or HiGrad branching \citep{Chen2020,FangXuYang2018,SuZhu2023}.

\subsection{Theoretical results for the geometric models}\label{app:geometric-results}
\Cref{sec:applications} gives the objectives, gradients, Hessians, and noise covariances for the geometric examples. The results below collect their sufficient conditions. The i.i.d. stream is independent of initialization.

\subsubsection{Smooth Riemannian \texorpdfstring{$M$}{M}-estimation}\label{sec:m-estimation}
\begin{proposition}[Riemannian $M$-estimation]\label{prop:M-estimator}
Suppose $x_\star$ is an isolated minimizer, $H$ in \eqref{eq:M-H-Sigma} is positive definite, and the following conditions hold on a normal neighborhood of $x_\star$: the mean gradient has the expansion \eqref{eq:mean-field-holder}; $\E\norm{\grad\ell(x_\star;Z)}^2<\infty$; the stochastic gradient satisfies \eqref{eq:noise-continuity}; and the stochastic gradient path satisfies localization and \eqref{eq:l2-rate}. Then, for $\beta$ satisfying \eqref{eq:beta-range},
\begin{equation}\label{eq:M-clt}
 \sqrt{N_n}\Log_{x_\star}(\widehat x_n)
 \rightsquig\cN(0,H^{-1}\Sigma H^{-1}).
\end{equation}
If $\Sigma$ is positive definite, \eqref{eq:confidence-region} has asymptotic coverage $1-\alpha$. More generally, every smooth $q$-dimensional functional satisfying the rank condition in \Cref{cor:functionals} admits the pivot in \eqref{eq:functional-pivot}.
\end{proposition}

\subsubsection{Intrinsic Fr\'echet means}
Write $\widehat\mu_n$ for the fixed-chart average $\widehat x_n$ when estimating the intrinsic mean $\mu$.
\begin{proposition}[Intrinsic Fr\'echet mean]\label{prop:frechet}
Suppose either of the following sets of sufficient conditions holds.
\begin{enumerate}[label=(\alph*),leftmargin=2.1em]
\item The support of $Z$ and the retained path lie in a compact strongly convex set separated by a positive distance from the relevant cut loci, and the Hessian of squared distance has a square integrable local H\"older envelope.
\item The manifold is Hadamard, $\E\dist(x_0,Z)^2<\infty$ for some $x_0$, and the local mean field and noise-continuity conditions hold near $\mu$.
\end{enumerate}
Assume $H_\mu$ is positive definite, the path satisfies localization and \eqref{eq:l2-rate}, and the step-size exponent satisfies \eqref{eq:beta-range} for the local mean field exponent $\rho$. Then
\begin{equation}\label{eq:frechet-clt}
 \sqrt{N_n}\Log_\mu(\widehat\mu_n)
 \rightsquig
 \cN(0,H_\mu^{-1}\Sigma_\mu H_\mu^{-1}).
\end{equation}
If $\Sigma_\mu$ is positive definite, \Cref{alg:one-path} gives an asymptotic $(1-\alpha)$ confidence region for $\mu$ without estimating either operator in \eqref{eq:frechet-H-Sigma}.
\end{proposition}

The support condition in part (a) gives a regular model with support separated from the cut locus. Refined central limit theorems permit support meeting the cut locus under geometry and local-density conditions \citep{HotzLeWood2024}. Smeary means have different scaling \citep{EltznerHuckemann2019}; empirical likelihood is a batch-sample alternative \citep{KurisuOtsu2026}.
For affine-invariant matrix means, \eqref{eq:frechet-clt} applies with $\mu=M$, $H_\mu=H_M$, and $\Sigma_\mu=\Sigma_M$.

\subsubsection{Streaming principal subspaces}
\begin{proposition}[Grassmann principal subspace]\label{prop:pca}
Assume $\lambda_r>\lambda_{r+1}$ and $\E\norm{X}^4<\infty$. Suppose the stochastic Grassmann recursion is localized in the basin of $[U_\star]$ and satisfies \eqref{eq:l2-rate}. Let $[\widehat U_n]$ be its fixed-chart average, and represent $H_{\mathrm{PCA}}$ in the column-stacked $K$ coordinates of \Cref{sec:pca}. For $1/2<\beta<1$,
\begin{equation}\label{eq:pca-clt}
 \sqrt{N_n}\,
 \vecop\{U_\perp^\transpose\Log_{[U_\star]}([\widehat U_n])\}
 \rightsquig
 \cN\left(0,
 H_{\mathrm{PCA}}^{-1}\Sigma_{\mathrm{PCA}}H_{\mathrm{PCA}}^{-1}
 \right).
\end{equation}
If $\Sigma_{\mathrm{PCA}}$ is positive definite on the tangent space, \Cref{alg:one-path} gives a full confidence region for the principal subspace. Otherwise, \Cref{cor:functionals} applies to every nondegenerate smooth contrast.
\end{proposition}

\subsection{Extensions}\label{sec:extensions-retraction}

The first order argument allows any uniformly $C^2$ local inverse retraction with nonsingular derivative on a common retained localization neighborhood, which can reduce computation required for repeated logarithm evaluations. A quotient-manifold extension additionally needs a locally smooth horizontal representation; singular orbits require critical values adapted to changing dimension or isotropy. Dependence is also possible in principle when the leading noise admits a martingale or Poisson-equation decomposition with a Brownian partial sum limit. Long-range dependence would change that limit and hence the critical values.

Further directions include rank adaptation and nonregular inference. Unknown covariance rank, increasing tangent dimension, singular Hessians, smeary intrinsic means, nonsmooth population objectives, and optional stopping all interact with either the convergence rate or the normalizer spectrum.  These developments would extend the present fixed-dimensional regular theory.

\section{Technical Lemmas}\label{app:technical}
\subsection{Local coordinate expansions}\label{app:local}

This appendix develops the geometric and stochastic approximation expansions used in the proofs. Write $b=b_n$ and $N=N_n$ throughout. Constants may change from line to line. We first construct local coefficients on the compact normal-coordinate set corresponding to $\overline U_0$, extend those coefficients to a globally defined burn-in-reset recursion, and only then transfer its limits to the original retained path on a high-probability agreement event.

\begin{lemma}[Differential of the logarithm map]\label{lem:dlog}
There are constants $r_0,C>0$ such that, whenever $x=\Exp_{x_\star}(\Delta)$ and $\norm{\Delta}\le r_0$,
\begin{equation}\label{eq:dlog-bound}
 \norm{D\Log_{x_\star}|_x-\mathsf P_{x\to x_\star}}_{\op}
 \le C\norm{\Delta}^2.
\end{equation}
\end{lemma}

\begin{proof}
Let $\gamma(t)=\Exp_{x_\star}(t\Delta)$ and use a parallel orthonormal frame along $\gamma$. For $w\in T_{x_\star}\M$, the differential $D\Exp_{x_\star}|_\Delta[w]$ is the value at $t=1$ of the Jacobi field $J$ satisfying $J(0)=0$ and $\nabla_tJ(0)=w$. After the path is rescaled to unit time, its coordinate vector $j(t)$ satisfies
\[
 j''(t)+\mathcal R_\Delta(t)j(t)=0,
 \qquad j(0)=0,
 \qquad j'(0)=w,
\]
where $\sup_t\norm{\mathcal R_\Delta(t)}_{\op}\le C\norm{\Delta}^2$ by bounded sectional curvature on the local compact set. Twice integrating the equation gives
\[
 j(1)=w-\int_0^1(1-s)\mathcal R_\Delta(s)j(s)\,ds.
\]
A Gronwall bound yields $\sup_{s\le1}\norm{j(s)}\le C\norm{w}$ and hence
$\norm{j(1)-w}\le C\norm{\Delta}^2\norm{w}$. Therefore
\[
 D\Exp_{x_\star}|_\Delta
 =\mathsf P_{x_\star\to x}\{\Id+E_\Delta\},
 \qquad \norm{E_\Delta}_{\op}\le C\norm{\Delta}^2.
\]
For $r_0$ small, $\Id+E_\Delta$ is invertible. Inverting this display proves \eqref{eq:dlog-bound}.
\end{proof}

\begin{lemma}[Small retained increments]\label{lem:small-step}
Under Assumptions \ref{ass:geometry} and \ref{ass:oracle}, and with $\beta>1/2$,
\begin{equation}\label{eq:small-step}
 \Pp\left(
 \max_{b_n\le k\le n-1}
 \eta_{k+1}\norm{G_{k+1}(x_k)}_{x_k}>r_{\mathcal R}/2,
 x_k\in U_0\text{ for all }b_n\le k\le n
 \right)\longrightarrow0.
\end{equation}
\end{lemma}

\begin{proof}
On the localization event, conditional Markov's inequality and
\eqref{eq:gradient-second-moment} give
\begin{align*}
 &\Pp\left(
 \max_{b_n\le k\le n-1}
 \eta_{k+1}\norm{G_{k+1}(x_k)}>r_{\mathcal R}/2,
 x_k\in U_0\text{ for all }b_n\le k\le n
 \right)\\
 &\qquad\le
 4r_{\mathcal R}^{-2}
 \sum_{k=b_n}^{n-1}\eta_{k+1}^2
 \E\left[
 \norm{G_{k+1}(x_k)}^2
 \ind\{x_k\in U_0\}
 \right]\\
 &\qquad\le C\sum_{k=b_n}^{\infty}k^{-2\beta}
 \longrightarrow0,
\end{align*}
where the last step uses $b_n\to\infty$ and $\beta>1/2$.
\end{proof}

\begin{lemma}[Local coefficient construction]\label{lem:local-recursion}
Under Assumptions \ref{ass:geometry}--\ref{ass:oracle}, whenever $x_k\in\overline U_0$, the increment-truncated update in the proof has normal-coordinate error $\Delta_k=\Log_{x_\star}(x_k)$ satisfying
\begin{equation}\label{eq:local-recursion}
 \Delta_{k+1}
 =\Delta_k-\eta_{k+1}(H\Delta_k+\xi_{k+1})
 +\eta_{k+1}a_{k+1}
 +\eta_{k+1}m_{k+1}
 +\eta_{k+1}^2q_{k+1},
\end{equation}
where $a_{k+1}$ is $\F_k$-measurable, $m_{k+1}$ is a martingale difference, and
\begin{align}
 \norm{a_{k+1}}
 &\le C\{\norm{\Delta_k}^{1+\rho}+\norm{\Delta_k}^2\},
 \label{eq:a-bound}\\
 \E(\norm{m_{k+1}}^2\mid\F_k)
 &\le C\{\norm{\Delta_k}^{2\bar\nu}+\norm{\Delta_k}^2\},
 \qquad \bar\nu=\min(\nu,1),
 \label{eq:m-bound}\\
 \E(\norm{q_{k+1}}\mid\F_k)&\le C.
 \label{eq:q-bound}
\end{align}
\end{lemma}

\begin{proof}
Let $\phi=\Log_{x_\star}$ on $U_1$ and, for $x\in\overline U_0$, define
$\Phi_x(v)=\phi\{\mathcal R_x(v)\}$ for $\norm v_x\le r_{\mathcal R}$.
Choose a smooth cutoff $\chi$ that is one on the ball of radius
$r_{\mathcal R}/2$ and zero outside the ball of radius $r_{\mathcal R}$, and set
\[
 \widetilde\Phi_x(v)
 =\phi(x)+\chi(v)\{\Phi_x(v)-\phi(x)\}
\]
on the retraction tube, extended by $\phi(x)$ outside the tube. Compactness and the
$C^2$ assumption imply a uniform second order expansion
\begin{equation}\label{eq:retraction-taylor}
 \widetilde\Phi_x(-\eta g)
 =\phi(x)-\eta D\phi_x[g]+\eta^2R(x,g,\eta),
 \qquad
 \norm{R(x,g,\eta)}\le C\norm g_x^2.
\end{equation}
By the probability statement \eqref{eq:small-step} in \Cref{lem:small-step}, this truncated expansion agrees with the original
recursion for every retained update with probability tending to one. It is therefore
enough to analyze the truncated local recursion.

Write $x=x_k$. The conditional
second moment bound in \eqref{eq:gradient-second-moment} and
\eqref{eq:retraction-taylor} give a term $q_{k+1}$ satisfying
\eqref{eq:q-bound}. For the conditional mean, Assumption \ref{ass:objective}
and \Cref{lem:dlog} yield
\begin{align}
 D\phi_x[\grad F(x)]
 &=\mathsf P_{x\to x_\star}\grad F(x)
   +\{D\phi_x-\mathsf P_{x\to x_\star}\}\grad F(x)\notag\\
 &=H\Delta_k+r_F(x),
 \qquad
 \norm{r_F(x)}\le
 C\{\norm{\Delta_k}^{1+\rho}+\norm{\Delta_k}^2\}.
 \label{eq:mean-field-expansion}
\end{align}
Expansion \eqref{eq:mean-field-expansion} also uses that $\grad F$ is bounded on the local compact set.

For the centered noise, put
\[
 m_{k+1}
 =-\left[D\phi_x\{\epsilon_{k+1}(x)\}-\xi_{k+1}\right].
\]
It is a martingale difference. By adding and subtracting
$\mathsf P_{x\to x_\star}\epsilon_{k+1}(x)$, using
\eqref{eq:noise-continuity}, \Cref{lem:dlog}, and the conditional second moment
bound, we obtain
\[
 \E(\norm{m_{k+1}}^2\mid\F_k)
 \le C\{\norm{\Delta_k}^{2\bar\nu}+\norm{\Delta_k}^2\},
 \qquad \bar\nu=\min(\nu,1).
\]
Finally set $a_{k+1}=-r_F(x_k)$, with its sign absorbed into the notation in
\eqref{eq:local-recursion}, and collect the Taylor remainder in $q_{k+1}$.
This proves \eqref{eq:local-recursion}--\eqref{eq:q-bound}.
\end{proof}

\subsubsection{Burn-in-reset localized recursion}\label{app:localization}
Let $\phi=\Log_{x_\star}$ and $\mathcal K=\phi(\overline U_0)$. For $\delta\in\mathcal K$, write $x(\delta)=\phi^{-1}(\delta)$ and let $\widetilde\Phi_x$ be the increment-truncated map in \eqref{eq:retraction-taylor}. Define
\begin{align*}
 r_F(\delta)
 &=D\phi_{x(\delta)}[\grad F\{x(\delta)\}]-H\delta,\\
 a^{\rm loc}_{k+1}(\delta)
 &=-\ind\{\delta\in\mathcal K\}r_F(\delta),\\
 m^{\rm loc}_{k+1}(\delta)
 &=\ind\{\delta\in\mathcal K\}
 \left[\xi_{k+1}
 -D\phi_{x(\delta)}\{\epsilon_{k+1}(x(\delta))\}\right],\\
 q^{\rm loc}_{k+1}(\delta)
 &=\ind\{\delta\in\mathcal K\}\eta_{k+1}^{-2}
 \left[
 \widetilde\Phi_{x(\delta)}
 \{-\eta_{k+1}G_{k+1}(x(\delta))\}
 -\delta
 +\eta_{k+1}D\phi_{x(\delta)}
 \{G_{k+1}(x(\delta))\}
 \right],
\end{align*}
where every coefficient is set to zero when $\delta\notin\mathcal K$. The state indicator is predictable. Assumption \ref{ass:oracle}, applied at predictable random arguments, therefore makes $m^{\rm loc}_{k+1}(\delta_k)$ a martingale difference for every $\F_k$-measurable $\delta_k$. The bounds \eqref{eq:a-bound}--\eqref{eq:m-bound} hold globally with $\Delta_k$ replaced by $\delta_k$. Moreover,
\[
 \norm{q^{\rm loc}_{k+1}(\delta_k)}
 \le C\ind\{\delta_k\in\mathcal K\}
 \norm{G_{k+1}\{x(\delta_k)\}}^2.
\]
For increments inside the inner cutoff tube this is \eqref{eq:retraction-taylor}. Outside that tube, $\eta_{k+1}\norm G>r_{\mathcal R}/2$ implies
$\eta_{k+1}\norm G\le (2/r_{\mathcal R})\eta_{k+1}^2\norm G^2$, which bounds both the omitted first order term and the truncated displacement. Hence
$\E(\norm{q^{\rm loc}_{k+1}(\delta_k)}\mid\F_k)\le C$ uses only \eqref{eq:gradient-second-moment}.

For each $n$, initialize
\[
 \Delta_{b_n,n}^{\rm loc}
 =\begin{cases}
 \Log_{x_\star}(x_{b_n}),&x_{b_n}\in U_0,\\
 0,&x_{b_n}\notin U_0.
 \end{cases}
\]
and, for $b_n\le k<n$, define the globally valid auxiliary recursion
\begin{equation}\label{eq:global-local-recursion}
\begin{split}
 \Delta_{k+1,n}^{\rm loc}
 ={}&\Delta_{k,n}^{\rm loc}
 -\eta_{k+1}(H\Delta_{k,n}^{\rm loc}+\xi_{k+1})\\
 &+\eta_{k+1}a^{\rm loc}_{k+1}(\Delta_{k,n}^{\rm loc})
 +\eta_{k+1}m^{\rm loc}_{k+1}(\Delta_{k,n}^{\rm loc})
 +\eta_{k+1}^2q^{\rm loc}_{k+1}(\Delta_{k,n}^{\rm loc}).
\end{split}
\end{equation}
Outside $\mathcal K$, this is the stable leading linear recursion driven by $\xi_{k+1}$. Assumption \ref{ass:stability} imposes the uniform mean square bound on the full triangular array in \eqref{eq:global-local-recursion}.

\begin{lemma}[Localization agreement and transfer]\label{lem:localization-transfer}
Let
\[
 E_n
 =\{x_k\in U_0\text{ for all }b_n\le k\le n\}
 \cap
 \left\{
 \max_{b_n\le k<n}
 \eta_{k+1}\norm{G_{k+1}(x_k)}
 \le r_{\mathcal R}/2
 \right\}.
\]
Then $\Pp(E_n)\to1$, and on $E_n$,
\[
 \Delta_{k,n}^{\rm loc}=\Log_{x_\star}(x_k),
 \qquad b_n\le k\le n.
\]
Consequently, any endpoint, $L^2$-path, estimator, normalizer, or statistic built from the original and auxiliary retained paths and agreeing on $E_n$ has the same weak limit whenever one of the two versions has a weak limit.
\end{lemma}

\begin{proof}
Assumption \ref{ass:geometry} and \Cref{lem:small-step} give $\Pp(E_n)\to1$. On $E_n$, the state indicator equals one and the increment-truncated map equals the original retraction at every retained update. Equality follows by induction from the common burn-in state. For any two corresponding random elements $X_n$ and $X_n^{\rm loc}$ in a metric space, $\Pp\{d(X_n,X_n^{\rm loc})>0\}\le\Pp(E_n^c)\to0$; Slutsky's theorem gives the transfer statement.
\end{proof}

\begin{lemma}[Expansion of the anchor logarithm]\label{lem:chart-expansion}
Let $a\in U_0$ and define
\[
 \Psi_a(x)=\mathsf P_{a\to x_\star}\Log_a(x)\in T_{x_\star}\M.
\]
There is a uniformly bounded invertible linear map $J_a:T_{x_\star}\M\to T_{x_\star}\M$ such that
\begin{equation}\label{eq:anchor-expansion}
 \Psi_a(x)-\Psi_a(x_\star)
 =J_a\Log_{x_\star}(x)+R_a(x),
 \qquad
 \norm{R_a(x)}\le C\dist(x,x_\star)^2.
\end{equation}
If $a\to x_\star$, then $J_a\to\Id$ in operator norm.
\end{lemma}

\begin{proof}
For $a$ and $x$ in the compact strongly convex set, $(a,x)\mapsto\Psi_a(x)$ is $C^2$. Apply Taylor's theorem to $x\mapsto\Psi_a(x)$ at $x_\star$ in normal coordinates. Its derivative is
\[
 J_a=D(\Psi_a\circ\Exp_{x_\star})_0,
\]
and the uniform bound on the second derivative gives the quadratic remainder in \eqref{eq:anchor-expansion}. The inverse function theorem shows that $J_a$ is invertible after $U_0$ is chosen sufficiently small; compactness makes the inverse bound uniform. At $a=x_\star$, $\Psi_{x_\star}(x)=\Log_{x_\star}(x)$ and hence $J_{x_\star}=\Id$. Continuity in $a$ gives the last assertion.
\end{proof}

\subsection{Averaging remainder}
We first isolate the averaging remainder for the globally defined auxiliary array. Within this subsection only, abbreviate $\Delta_{k,n}^{\rm loc}$ by $\Delta_k$ and the coefficients in \eqref{eq:global-local-recursion} by $a_{k+1}$, $m_{k+1}$, and $q_{k+1}$. All norms in this subsection are in $T_{x_\star}\M$.

\begin{lemma}[Abel decomposition]\label{lem:abel}
Under the recursion \eqref{eq:global-local-recursion}, for $1\le m\le N$,
\begin{equation}\label{eq:abel-decomposition}
 H\sum_{k=b}^{b+m-1}\Delta_k
 =-\sum_{k=b}^{b+m-1}\xi_{k+1}+R_{m,n},
\end{equation}
where $R_{m,n}$ is the sum of
\begin{align}
 R^{(1)}_{m,n}
 &=\eta_{b+1}^{-1}\Delta_b
 -\eta_{b+m}^{-1}\Delta_{b+m},
 \label{eq:R1}\\
 R^{(2)}_{m,n}
 &=\sum_{k=b+1}^{b+m-1}
 (\eta_{k+1}^{-1}-\eta_k^{-1})\Delta_k,
 \label{eq:R2}\\
 R^{(3)}_{m,n}
 &=\sum_{k=b}^{b+m-1}a_{k+1},
 \qquad
 R^{(4)}_{m,n}=\sum_{k=b}^{b+m-1}m_{k+1},
 \label{eq:R34}\\
 R^{(5)}_{m,n}
 &=\sum_{k=b}^{b+m-1}\eta_{k+1}q_{k+1}.
 \label{eq:R5}
\end{align}
The groups in \eqref{eq:R1}, \eqref{eq:R2}, \eqref{eq:R34}, and \eqref{eq:R5} exhaust $R_{m,n}$. A harmless shift of the first or last iterate changes \eqref{eq:abel-decomposition} by $o_p(\sqrt N)$.
\end{lemma}

\begin{proof}
Divide \eqref{eq:global-local-recursion} by $\eta_{k+1}$ and rearrange:
\[
 H\Delta_k
 =\eta_{k+1}^{-1}(\Delta_k-\Delta_{k+1})
 -\xi_{k+1}+a_{k+1}+m_{k+1}+\eta_{k+1}q_{k+1}.
\]
Summation from $k=b$ to $b+m-1$ and discrete integration by parts give \eqref{eq:R1}--\eqref{eq:R5}.
\end{proof}

\begin{lemma}[Negligibility of the Abel remainder]\label{lem:remainder}
Under Assumptions \ref{ass:geometry}--\ref{ass:stability},
\begin{equation}\label{eq:remainder-endpoint}
 \frac{1}{\sqrt N}R_{N,n}\xrightarrow{p}0
\end{equation}
and
\begin{equation}\label{eq:remainder-L2}
 \frac{1}{N^2}\sum_{m=1}^{N}\norm{R_{m,n}}^2\xrightarrow{p}0.
\end{equation}
\end{lemma}

\begin{proof}
We treat the five terms separately.

For $R^{(1)}_{m,n}$, Assumption \ref{ass:stability} gives
\[
 \E\norm{\eta_k^{-1}\Delta_k}^2
 \le C\eta_k^{-1}\le Ck^\beta.
\]
Therefore
\[
 \frac1N\E\norm{R^{(1)}_{N,n}}^2
 \le C\frac{b^\beta+n^\beta}{N}\longrightarrow0,
\]
because $b/n\to0$ and $\beta<1$. Also,
\[
 \frac{1}{N^2}\sum_{m=1}^{N}
 \E\norm{\eta_{b+m}^{-1}\Delta_{b+m}}^2
 \le\frac{C}{N^2}\sum_{k=b+1}^{n}k^\beta
 =O(n^{\beta-1})\to0.
\]
The initial boundary contributes $N^{-1}\eta_b^{-1}=O(b^\beta/N)\to0$ to the same display.

For $R^{(2)}_{m,n}$, the mean value theorem gives
\[
 0\le\eta_{k+1}^{-1}-\eta_k^{-1}\le Ck^{\beta-1}.
\]
Minkowski's inequality and \eqref{eq:l2-rate} yield
\begin{align*}
 \{\E\norm{R^{(2)}_{m,n}}^2\}^{1/2}
 &\le C\sum_{k=b+1}^{b+m-1}k^{\beta-1}k^{-\beta/2}\\
 &\le C(b+m)^{\beta/2}.
\end{align*}
Consequently,
\[
 \frac1N\E\norm{R^{(2)}_{N,n}}^2=O(n^{\beta-1})\to0,
 \qquad
 \frac1{N^2}\sum_{m=1}^{N}\E\norm{R^{(2)}_{m,n}}^2
 =O(n^{\beta-1})\to0.
\]

For $R^{(3)}_{m,n}$, \eqref{eq:a-bound}, Jensen's inequality, and local boundedness give
\begin{align*}
 \E\norm{a_{k+1}}
 &\le C\left\{
 (\E\norm{\Delta_k}^2)^{(1+\rho)/2}
 +\E\norm{\Delta_k}^2
 \right\}\\
 &\le C\{k^{-\beta(1+\rho)/2}+k^{-\beta}\}.
\end{align*}
Hence
\[
 \frac1{\sqrt N}\E\max_{m\le N}\norm{R^{(3)}_{m,n}}
 \le\frac1{\sqrt N}\sum_{k=b}^{n-1}\E\norm{a_{k+1}}
 =o(1)
\]
by $\beta(1+\rho)>1$ and $\beta>1/2$. Markov's inequality proves uniform $o_p(\sqrt N)$ control, which implies both \eqref{eq:remainder-endpoint} and \eqref{eq:remainder-L2} for this term.

For $R^{(4)}_{m,n}$, Doob's inequality and \eqref{eq:m-bound} imply
\begin{align*}
 \frac1N\E\max_{m\le N}\norm{R^{(4)}_{m,n}}^2
 &\le\frac{C}{N}\sum_{k=b}^{n-1}
 \E\{\norm{\Delta_k}^{2\bar\nu}+\norm{\Delta_k}^2\}\\
 &\le\frac{C}{N}\sum_{k=b}^{n-1}
 \{\eta_k^{\bar\nu}+\eta_k\}
 \longrightarrow0.
\end{align*}
For $\bar\nu\le1$, the second line uses Jensen's inequality; local boundedness covers any original $\nu>1$. Thus this term is uniformly $o_p(\sqrt N)$.

Finally, \eqref{eq:q-bound} gives
\[
 \frac1{\sqrt N}
 \E\max_{m\le N}\norm{R^{(5)}_{m,n}}
 \le\frac{C}{\sqrt N}\sum_{k=b}^{n-1}\eta_{k+1}
 =O(n^{1/2-\beta})\to0.
\]
Combining the five estimates proves the lemma.
\end{proof}

The same estimates quantify the auxiliary averaging error:
\[
 \frac{\norm{R_{N,n}}}{\sqrt N}
 +\left\{\frac1{N^2}\sum_{m=1}^N\norm{R_{m,n}}^2\right\}^{1/2}
 =O_p\!\left(n^{-(1-\beta)/2}
 +n^{-\{\beta(1+\rho)-1\}/2}
 +n^{1/2-\beta}+n^{-\beta\bar\nu/2}\right).
\]
The first term comes from boundaries and changing gains, the second from the mean-field remainder, the third from quadratic and retraction terms, and the fourth from state-dependent noise. For $\rho=1$, choosing $\beta=0.6$ gives a slowest bound $n^{-0.1}$ when $\bar\nu\ge1/3$; taking $\beta$ near one instead slows the boundary bound. These are stochastic-approximation remainder bounds, not coverage-error rates: the martingale approximation, localization probability, and inverse-normalizer tails require separate quantitative control.

\subsection{Chart and bridge limits}
Here $\Delta_k$ again denotes the original normal-coordinate path defined in \Cref{sec:assumptions}; $\Delta_{k,n}^{\rm loc}$ continues to denote the auxiliary array.

\begin{lemma}[Chart partial sums]\label{lem:chart-partial}
Let
\[
 v^\star_n=\Log_{a_n}(x_\star),
 \qquad
 Y_{k,n}=\mathsf P_{a_n\to x_\star}(v_{k,n}-v^\star_n).
\]
Under the assumptions of \Cref{thm:selfnorm},
\begin{equation}\label{eq:Y-linearization}
 \frac1{\sqrt N}
 \sum_{k=b+1}^{b+\lfloor Nr\rfloor}
 \{Y_{k,n}-J_{a_n}\Delta_k\}
 \xrightarrow{p}0
\end{equation}
uniformly in $r\in[0,1]$, and $J_{a_n}\xrightarrow{p}\Id$.
\end{lemma}

\begin{proof}
On the agreement event $E_n$ in \Cref{lem:localization-transfer}, expansion
\eqref{eq:anchor-expansion} gives
\[
 \norm{Y_{k,n}-J_{a_n}\Delta_k}
 \le C\norm{\Delta_{k,n}^{\rm loc}}^2.
\]
Consequently, for every $\epsilon>0$,
\begin{align*}
 &\Pp\left\{
 \frac1{\sqrt N}\sum_{k=b+1}^{n}
 \norm{Y_{k,n}-J_{a_n}\Delta_k}>\epsilon
 \right\}\\
 &\qquad\le \Pp(E_n^c)
 +\frac{C}{\epsilon\sqrt N}
 \sum_{k=b+1}^{n}\E\norm{\Delta_{k,n}^{\rm loc}}^2\\
 \le \Pp(E_n^c)+\frac{C}{\epsilon\sqrt N}
 \sum_{k=b+1}^{n}\eta_k\longrightarrow0.
\end{align*}
This bound controls every partial sum and proves the uniform convergence in
\eqref{eq:Y-linearization}. Moreover,
\[
 \Pp\{\dist(a_n,x_\star)>\epsilon\}
 \le \Pp(E_n^c)
 +\epsilon^{-2}\E\norm{\Delta_{b_n,n}^{\rm loc}}^2
 \longrightarrow0.
\]
Thus $a_n\to x_\star$ in probability, and the last part of
\Cref{lem:chart-expansion} gives $J_{a_n}\to\Id$.
\end{proof}

\begin{lemma}[High-probability inverse-chart domain]\label{lem:chart-domain}
There is an open
neighborhood $\mathcal D$ of $(x_\star,0)$ in the tangent bundle on which
$(a,v)\mapsto\Exp_a(v)$ takes values in the common normal neighborhood
$U_1$, and
\[
 \Pp\{(a_n,\bar v_n)\in\mathcal D\}\longrightarrow1.
\]
For any coordinate family $\chi_a$ satisfying
\Cref{prop:coordinate-reduction}(b), the analogous statement holds for a
common domain of $(a,z)\mapsto\chi_a^{-1}(z)$, with
$\bar z_n=N_n^{-1}\sum_{k=b_n+1}^n\chi_{a_n}(x_k)$.
\end{lemma}

\begin{proof}
The proof of \Cref{lem:chart-partial} gives $a_n\to x_\star$ in probability.
Combining the endpoint case of \Cref{lem:chart-partial} with
\Cref{thm:integrated-fclt} gives
$\bar v_n-v_n^\star=O_p(N_n^{-1/2})$. Smoothness of the logarithm on the
common normal neighborhood gives $v_n^\star\to0$ in probability. Hence
$(a_n,\bar v_n)\to(x_\star,0)$ in probability. Local smoothness and the
inverse function theorem provide an open neighborhood $\mathcal D$ of that
limit on which the exponential chart maps into $U_1$, proving the first
claim.

For the alternative chart, the uniform expansion
\eqref{eq:alternative-chart-expansion}, the same event-splitting argument as
in \Cref{lem:chart-partial}, and the endpoint partial sum limit give
$\bar z_n-\chi_{a_n}(x_\star)=O_p(N_n^{-1/2})$. Joint continuity gives
$\chi_{a_n}(x_\star)\to0$. The parameterized inverse function theorem then
provides a common open inverse-chart domain containing $(a_n,\bar z_n)$ with
probability tending to one.
\end{proof}

\begin{lemma}[Positive definiteness of the Brownian bridge Gram matrix]\label{lem:Q-positive}
For fixed $d$, the matrix $Q_d$ in \eqref{eq:pivot-limit} is positive definite almost surely.
\end{lemma}

\begin{proof}
Choose distinct points $0<t_1<\cdots<t_d<1$ and form the random matrix
$M=((B_d(t_j))_i)_{j,i=1}^d$. Its columns are independent, each with positive definite covariance
$(\min(t_j,t_\ell)-t_jt_\ell)_{j,\ell}$. Hence $\vecop(M)$ is a nondegenerate Gaussian vector in $\R^{d^2}$, so $M$ has a Lebesgue density. The polynomial $\det(M)$ is not identically zero; therefore $\Pp\{\det(M)=0\}=0$.

If $Q_d$ were singular, there would be a nonzero $u\in\R^d$ such that
\[
 0=u^\transpose Q_du=\int_0^1\{u^\transpose B_d(r)\}^2dr.
\]
Continuity of $B_d$ would imply $u^\transpose B_d(r)=0$ for all $r$, and hence $Mu=0$. Thus $M$ would be singular. The preceding paragraph proves that this event has probability zero.
\end{proof}

\section{Proofs}\label{app:proofs}
Write $b=b_n$ and $N=N_n$ throughout this appendix.

\subsection{Proof of Proposition~\ref{prop:stability}}\label{app:stability}

\begin{proof}[Proof of \Cref{prop:stability}]
We first verify the sufficient condition for the drift. Write $\phi=\Log_{x_\star}$, $\delta=\Delta_{k,n}^{\rm loc}$, and $\eta=\eta_{k+1}$. For $\delta\in\mathcal K=\phi(\overline U_0)$, the coefficient construction in \Cref{lem:local-recursion} gives exactly
\[
 \Delta_{k+1,n}^{\rm loc}=\widetilde\Phi_x\{-\eta G_{k+1}(x)\},\qquad x=\phi^{-1}(\delta).
\]
The cutoff map has uniformly bounded first and second derivatives in its increment argument. In particular, its displacement from $\phi(x)$ is bounded by $C\eta\norm{G_{k+1}(x)}$ for every increment, including those outside the retraction tube. For $D=\Delta_{k+1,n}^{\rm loc}-\delta$, conditional second moments and \eqref{eq:retraction-taylor} therefore imply
\[
 \E(\norm D^2\mid\F_k)\le C\eta^2,\qquad
 \E(D\mid\F_k)=-\eta D\phi_x[\grad F(x)]+r_k,\quad \norm{r_k}\le C\eta^2.
\]
The Gauss lemma identifies the radial derivative exactly:
\[
 \ip{\delta}{D\phi_x[\grad F(x)]}
 =\ip{\delta}{\mathsf P_{x\to x_\star}\grad F(x)}
 \ge\mu_0\norm\delta^2.
\]
Expanding $\norm{\delta+D}^2$ and using boundedness of $\mathcal K$ gives
$\E(\norm{\Delta_{k+1,n}^{\rm loc}}^2\mid\F_k)
\le(1-2\mu_0\eta)\norm\delta^2+C\eta^2$.
For $\delta\notin\mathcal K$, the update is $(\Id-\eta H)\delta-\eta\xi_{k+1}$. Conditional centering and positive definiteness of $H$ give the same form of bound with a possibly smaller positive drift constant, for all sufficiently small $\eta$. Since $b_n\to\infty$, these bounds establish \eqref{eq:one-step-drift} uniformly over the retained array for all sufficiently large $n$, without using \eqref{eq:l2-rate}.

Let $u_{k,n}=\E\norm{\Delta_{k,n}^{\rm loc}}^2$. Taking expectations in \eqref{eq:one-step-drift} gives
\begin{equation*}
 u_{k+1,n}\le(1-c_0\eta_{k+1})u_{k,n}+C_0\eta_{k+1}^2.
\end{equation*}
For $\eta_k=\eta_0k^{-\beta}$ with $\beta<1$,
\[
 \frac{\eta_k}{\eta_{k+1}}
 =\left(1+\frac1k\right)^\beta
 \le1+\frac{C}{k}.
\]
Because $k^{-1}=o(\eta_{k+1})$, there is $k_0$ such that
$C/k\le c_0\eta_{k+1}/2$ for $k\ge k_0$. Choose $K\ge C_b$ so large that $C_0\le Kc_0/4$. If $u_{k,n}\le K\eta_k$, then
\begin{align*}
 u_{k+1,n}
 &\le K\eta_k(1-c_0\eta_{k+1})+C_0\eta_{k+1}^2\\
 &\le K\eta_{k+1}
 \left(1+\frac{c_0}{2}\eta_{k+1}\right)
 (1-c_0\eta_{k+1})
 +C_0\eta_{k+1}^2\\
 &\le K\eta_{k+1}
 -\frac{Kc_0}{4}\eta_{k+1}^2
 +C_0\eta_{k+1}^2
 \le K\eta_{k+1},
\end{align*}
where $k_0$ can be enlarged if necessary. Take $n$ sufficiently large that $b_n\ge k_0$. The uniform initial bound gives $u_{b_n,n}\le K\eta_{b_n}$, and induction from $b_n$ to $n$ proves \eqref{eq:l2-rate} with constants independent of $n$.

\end{proof}

\subsection{Proof of Theorem~\ref{thm:integrated-fclt}}\label{app:partial-sum}

\begin{proof}[Proof of \Cref{thm:integrated-fclt}]
In this proof, $\Delta_k$ abbreviates the auxiliary array $\Delta_{k,n}^{\rm loc}$; the last step transfers the limit to the original path.
By the martingale functional central limit theorem under Assumption \ref{ass:fclt},
\begin{equation}\label{eq:xi-fclt}
 \frac1{\sqrt N}\sum_{k=b}^{b+\lfloor Nr\rfloor-1}\xi_{k+1}
 \rightsquigarrow\Sigma^{1/2}W_d(r)
\end{equation}
in $D([0,1];\R^d)$. Since the limit is continuous, this also implies convergence in $L^2[0,1]$, jointly with the endpoint.

By \Cref{lem:abel,lem:remainder}, the difference between the retained $\Delta$ partial sum path and $-H^{-1}$ times the path in \eqref{eq:xi-fclt} converges to zero in $L^2$, and the endpoint difference is $o_p(1)$. For the shift between sums starting at $b$ and $b+1$, the path difference at $r$ is $N^{-1/2}(\Delta_{b+\lfloor Nr\rfloor}-\Delta_b)$. Therefore
\[
 \int_0^1
 \norm{\frac{\Delta_{b+\lfloor Nr\rfloor}-\Delta_b}{\sqrt N}}^2dr
 \le \frac{2}{N}\norm{\Delta_b}^2
 +\frac{2}{N^2}\sum_{k=b}^{n-1}\norm{\Delta_k}^2
 \xrightarrow{p}0
\]
by \eqref{eq:l2-rate} and Markov's inequality. Separately, the endpoint shift is bounded by $N^{-1/2}(\norm{\Delta_b}+\norm{\Delta_n})=o_p(1)$. Slutsky's theorem in $L^2\times\R^d$ proves the auxiliary version of \eqref{eq:joint-L2-limit}. The agreement event in \Cref{lem:localization-transfer} then transfers the joint $L^2$-path and endpoint limit to the retained original recursion, proving \eqref{eq:joint-L2-limit}.
\end{proof}

\subsection{Proof of Theorem~\ref{thm:selfnorm}}\label{app:selfnorm}
\begin{proof}[Proof of \Cref{thm:selfnorm}]
Here $\Delta_k$ denotes the original normal-coordinate path from Section~\ref{sec:assumptions}.
Write $b=b_n$ and $N=N_n$. Parallel transport all vectors and operators from $T_{a_n}\M$ to $T_{x_\star}\M$; parallel transport is an isometry and does not change the statistic.

By \Cref{lem:chart-partial,thm:integrated-fclt},
\begin{equation}\label{eq:chart-process-limit}
 \left
 \{
 \frac1{\sqrt N}\sum_{k=b+1}^{b+\lfloor Nr\rfloor}
 (v_{k,n}-v_n^\star),
 \frac1{\sqrt N}\sum_{k=b+1}^{n}(v_{k,n}-v_n^\star)
 \right\}
 \rightsquigarrow
 \{-LW_d(r),-LW_d(1)\},
\end{equation}
where $L=H^{-1}\Sigma^{1/2}$, and the equality is understood after transport. The sign agrees with \Cref{thm:integrated-fclt} and cancels in the quadratic statistic.

Because constants cancel from the bridge, \eqref{eq:chart-process-limit} gives, in $L^2[0,1]$,
\begin{equation}\label{eq:bridge-process-limit}
 \frac1{\sqrt N}C_{\lfloor Nr\rfloor,n}
 \rightsquigarrow -LB_d(r).
\end{equation}
To transfer \eqref{eq:bridge-process-limit} to the normalizer, note that the map $f\mapsto\int_0^1f(r)f(r)^\transpose dr$ is continuous from $L^2([0,1];\R^d)$ to the space of symmetric matrices: indeed,
\[
 \norm{\int(ff^\transpose-gg^\transpose)}_{\op}
 \le(\norm{f}_{L^2}+\norm{g}_{L^2})\norm{f-g}_{L^2}.
\]
Because $C_{0,n}=C_{N,n}=0$, the discrete sum in \eqref{eq:self-normalizer} is exactly the integral of the corresponding step process. Thus
\[
 \widehat V_n\rightsquig LQ_dL^\transpose.
\]
This proves part (b). If $\Sigma$ is positive definite, then $L$ is invertible; \Cref{lem:Q-positive} implies that the limiting normalizer is positive definite and hence that $\widehat V_n$ is invertible with probability tending to one.

For the point estimator, \eqref{eq:chart-process-limit} gives
\begin{equation}\label{eq:vbar-limit}
 \sqrt N\,\mathsf P_{a_n\to x_\star}(\bar v_n-v_n^\star)
 \rightsquigarrow -LW_d(1).
\end{equation}
By \Cref{lem:chart-domain}, the inverse chart is evaluated on its common
domain with probability tending to one. On that event, the inverse chart
expansion from \Cref{lem:chart-expansion} gives
\[
 \Log_{x_\star}\{\Exp_{a_n}(\bar v_n)\}
 =J_{a_n}^{-1}
 \mathsf P_{a_n\to x_\star}(\bar v_n-v_n^\star)
 +O_p(\norm{\bar v_n-v_n^\star}^2).
\]
The squared term is $O_p(N^{-1})$, and $J_{a_n}\to\Id$, so \eqref{eq:intrinsic-clt} follows.

At $x=x_\star$, the numerator in \eqref{eq:test-statistic}, after multiplication by $\sqrt N$ and transport, converges jointly with $\widehat V_n$ to $-LW_d(1)$. Therefore
\begin{align*}
 T_n(x_\star)
 &\rightsquig
 W_d(1)^\transpose L^\transpose
 (LQ_dL^\transpose)^{-1}
 LW_d(1)\\
 &=W_d(1)^\transpose Q_d^{-1}W_d(1)=\cT_d.
\end{align*}
A Brownian bridge is independent of $W_d(1)$. Conditional on the positive definite $Q_d$, the final quadratic form has a continuous distribution. Hence the unconditional distribution is continuous, and \eqref{eq:coverage} follows.
\end{proof}

\subsection{Proof of Corollary~\ref{cor:functionals}}
\begin{proof}[Proof of \Cref{cor:functionals}]
The intrinsic delta method applied to \eqref{eq:intrinsic-clt} gives \eqref{eq:functional-clt}. By consistency and continuity,
$D_n$ converges, after the natural tangent identifications, to $Dh_{x_\star}$. The numerator and the transformed self-normalizer therefore have a joint limit of the form
\[
 -A_hW_d(1),
 \qquad
 A_h Q_d A_h^\transpose,
\]
where $A_h$ has row rank $q$. Let $C_A=(A_hA_h^\transpose)^{1/2}$ and $\widetilde W_q=C_A^{-1}A_hW_d$. Then $\widetilde W_q$ is a standard $q$-dimensional Brownian motion and
\[
 A_h Q_d A_h^\transpose
 =C_A\left\{\int_0^1
 (\widetilde W_q(r)-r\widetilde W_q(1))
 (\widetilde W_q(r)-r\widetilde W_q(1))^\transpose\,dr\right\}C_A.
\]
By \Cref{lem:Q-positive}, this limit is positive definite almost surely, so $\widehat V_{h,n}$ is invertible with probability tending to one. Cancelling $C_A$ from the projected endpoint and bridge normalizer yields $\cT_q$.
\end{proof}

\begin{proof}[Proof of \Cref{cor:derivative-approximation}]
The convergence of $\widehat V_n$ in \eqref{eq:V-limit} implies
$\norm{\widehat V_n}_{\op}=O_p(1)$, while continuity of $Dh$ and consistency
of $(a_n,\bar v_n)$ imply $\norm{D_n}_{\op}=O_p(1)$. Hence
\begin{align*}
 \norm{\widehat V^{\rm app}_{h,n}-\widehat V_{h,n}}_{\op}
 &\le
 \norm{\widehat D_n-D_n}_{\op}\norm{\widehat V_n}_{\op}
 \{\norm{\widehat D_n}_{\op}+\norm{D_n}_{\op}\}
 =o_p(1).
\end{align*}
The projected endpoint and normalizer therefore have the same joint limit as
in the proof of \Cref{cor:functionals}. That limit is positive definite almost
surely under the row-rank condition, so continuity of matrix inversion and
Slutsky's theorem prove the first claim.

For the centered-difference statement, choose orthonormal coordinates in the
fixed-dimensional space $T_{a_n}\M$ and write
$g_n(v)=h\{\Exp_{a_n}(v)\}$. Componentwise Taylor expansion at $\bar v_n$
gives
\[
 \frac{g_n(\bar v_n+s_ne_j)-g_n(\bar v_n-s_ne_j)}{2s_n}
 =Dg_n(\bar v_n)[e_j]+O_p(s_n^2),
\]
uniformly over the finitely many coordinate directions. The common-domain
and uniform $C^3$ conditions make the remainder uniform, hence
$\norm{\widehat D_n-D_n}_{\op}=O_p(s_n^2)=o_p(1)$. This argument needs no
comparison between $s_n$ and $N_n^{-1/2}$.
\end{proof}

\begin{proof}[Proof of \Cref{prop:ridge-stability}]
If $\Sigma$ is positive definite, the joint endpoint and normalizer limit is
\[
 \{Z_n,\widetilde V_n,\lambda_n\}
 \rightsquig
 \{-LW_d(1),LQ_dL^\transpose,0\},
 \qquad L=H^{-1}\Sigma^{1/2},
\]
where $Z_n=\sqrt N\,\mathsf P_{a_n\to x_\star}(\bar v_n-v_n^\star)$ and $\widetilde V_n$ is the transported normalizer in \Cref{thm:selfnorm}. Its limiting normalizer is positive definite almost surely. Joint Slutsky convergence, which does not require independence of the ridge, gives
\[
 \{Z_n,\widetilde V_n+\lambda_n\Id\}
 \rightsquigarrow
 \{-LW_d(1),LQ_dL^\transpose\}.
\]
Matrix inversion is continuous on the SPD matrix cone, so the ridge
does not change the limiting quadratic form. Under the row-rank condition of \Cref{cor:functionals}, its proof gives the joint convergence of the functional numerator, projected normalizer, and ridge:
\[
 \left\{\sqrt N\{h(\widehat x_n)-h(x_\star)\},
 \widehat V_{h,n},\lambda_{h,n}\right\}
 \rightsquigarrow
 \{-A_hW_d(1),A_hQ_dA_h^\transpose,0\}.
\]
The limiting projected normalizer is positive definite almost surely, so the same Slutsky and continuous-mapping argument applies to $\widehat V_{h,n}+\lambda_{h,n}\Id_q$ and, by \Cref{cor:derivative-approximation}, to its derivative-approximated version.
\end{proof}

\subsection{Proof of Proposition~\ref{prop:coordinate-reduction}}\label{app:comparisons}

\begin{proof}[Proof of \Cref{prop:coordinate-reduction}]
For part (a), let $O$ be an orthogonal change of basis in $T_{a_n}\M$.
Then every $v_{k,n}$ and $C_{m,n}$ is replaced by $Ov_{k,n}$ and
$OC_{m,n}$, while $\widehat V_n$ is replaced by
$O\widehat V_nO^\transpose$. The quadratic form in
\eqref{eq:test-statistic} is therefore unchanged.

For part (b), transport the alternative coordinates to $T_{x_\star}\M$ and
write
\[
 \Xi_a(x)=\mathsf P_{a\to x_\star}\chi_a(x).
\]
On the event $E_n\cap\{a_n\in\mathcal A\}$, which has probability tending to one, every retained iterate lies in the common forward-chart domain. A uniform Taylor expansion at $x_\star$ therefore gives
\begin{equation}\label{eq:alternative-chart-expansion}
 \Xi_a(x)-\Xi_a(x_\star)
 =K_a\Log_{x_\star}(x)+R_a^\chi(x),
 \qquad
 \norm{R_a^\chi(x)}\le C\dist(x,x_\star)^2,
\end{equation}
where $K_a=D(\Xi_a\circ\Exp_{x_\star})_0$ and $K_a$ and $K_a^{-1}$
are uniformly bounded. Since $a_n\to x_\star$ in probability,
$K_{a_n}\to K_\star$ for a nonsingular $K_\star$. The proof of
\Cref{lem:chart-partial} applied to \eqref{eq:alternative-chart-expansion}
shows that the alternative-coordinate endpoint and bridge limits are multiplied by $K_\star$. Thus the limiting numerator and normalizer are
$-K_\star LW_d(1)$ and $K_\star LQ_dL^\transpose K_\star^\transpose$.
When $\Sigma$ is positive definite, the nonsingular factor $K_\star L$ cancels from the full quadratic form. By \Cref{lem:chart-domain}, the alternative inverse chart is evaluated on a common domain with probability tending to one. Writing $\bar z_n=N^{-1}\sum_{k=b+1}^n\chi_{a_n}(x_k)$, its Taylor expansion gives
\[
 \sqrt N\Log_{x_\star}\{\chi_{a_n}^{-1}(\bar z_n)\}
 =K_{a_n}^{-1}\mathsf P_{a_n\to x_\star}
 \sqrt N\{\bar z_n-\chi_{a_n}(x_\star)\}+o_p(1),
\]
and hence the intrinsic limit \eqref{eq:intrinsic-clt}, without requiring $\Sigma$ to be nonsingular.

For a smooth functional, the derivative of $h\circ\chi_{a_n}^{-1}$ at $\bar z_n$, after the natural tangent identifications, converges to $Dh_{x_\star}K_\star^{-1}$. Premultiplying the alternative-coordinate endpoint and bridge limits thus gives
\[
 -A_hW_d(1),
 \qquad
 A_hQ_dA_h^\transpose.
\]
The row-rank argument in the proof of \Cref{cor:functionals} yields the $\cT_q$ limit without inverting $L$. An operator-norm-consistent derivative approximation perturbs the tight alternative-coordinate normalizer by $o_p(1)$, exactly as in the proof of \Cref{cor:derivative-approximation}.

For part (c), in Euclidean space
$v_{k,n}=x_k-a_n$ and
$\widehat x_n=N_n^{-1}\sum_{k=b_n+1}^n x_k$. Moreover,
\[
 C_{m,n}
 =\sum_{j=1}^m x_{b_n+j}
 -\frac{m}{N_n}\sum_{j=1}^{N_n}x_{b_n+j},
\]
so the anchor cancels exactly. Equations \eqref{eq:self-normalizer} and
\eqref{eq:test-statistic} are consequently the Euclidean random scaling
normalizer and statistic after relabeling the retained observations.
\end{proof}

\subsection{Proof of Proposition~\ref{prop:wald-comparison}}
\begin{proof}[Proof of \Cref{prop:wald-comparison}]
Because $H$ and $\Sigma$ are positive definite,
$\Omega=H^{-1}\Sigma H^{-1}$ is positive definite. Equation
\eqref{eq:Omega-consistency}, together with eventual self-adjoint positive
definiteness of $\widehat\Omega_n$, implies after transport that
$\widehat\Omega_n^{-1}\xrightarrow{p}\Omega^{-1}$ in operator norm.
At $x=x_\star$, \eqref{eq:vbar-limit} and
\eqref{eq:Omega-consistency} imply, by Slutsky's theorem,
\[
 N_n\ip{\bar v_n-v_n^\star}
 {\widehat\Omega_n^{-1}(\bar v_n-v_n^\star)}_{a_n}
 \rightsquigarrow\chi_d^2.
\]
The self-normalized convergence is \eqref{eq:pivot-convergence}.

Ignoring the asymptotically inactive restriction to the common chart, the
image of the Wald region under $\Log_{a_n}$ is an ellipsoid with center
$\bar v_n$, shape matrix $\widehat\Omega_n$, and squared radius
$\chi^2_{d,1-\alpha}/N_n$. Hence its tangent space volume equals
\[
 \omega_d\left(\frac{\chi^2_{d,1-\alpha}}{N_n}\right)^{d/2}
 \det(\widehat\Omega_n)^{1/2}.
\]
Parallel transport preserves determinants in orthonormal frames, and
$\Omega=LL^\transpose$ with $L=H^{-1}\Sigma^{1/2}$. This proves
\eqref{eq:wald-volume}.

The image of the self-normalized region is similarly an ellipsoid with shape
matrix $\widehat V_n$ and squared radius $c_{d,1-\alpha}/N_n$. By
\eqref{eq:V-limit} and the continuous mapping theorem,
\[
 \det(\widehat V_n)^{1/2}
 \rightsquigarrow
 \det(LQ_dL^\transpose)^{1/2}
 =\abs{\det L}\det(Q_d)^{1/2},
\]
which proves \eqref{eq:sn-volume}. The two limits contain the same
model-dependent determinant but different pivotal factors, so no deterministic
volume ordering follows.
\end{proof}

\subsection{Verification for the applications}\label{app:applications}

\begin{proof}[Proof of \Cref{prop:M-estimator}]
Differentiation under the expectation gives conditional unbiasedness and the
operators in \eqref{eq:M-H-Sigma}. At the minimizer,
$\E\{\grad\ell(x_\star;Z)\}=0$, so the second operator is the covariance of
the stochastic gradient noise.

It remains to verify the local second moment condition. Let
$\epsilon(x;Z)=\grad\ell(x;Z)-\grad F(x)$. By local noise continuity,
\begin{align*}
 \E\norm{\epsilon(x;Z)}^2
 &\le
 2\E\norm{\mathsf P_{x\to x_\star}\epsilon(x;Z)-\epsilon(x_\star;Z)}^2
 +2\E\norm{\epsilon(x_\star;Z)}^2\\
 &\le C.
\end{align*}
Thus the noise has a locally bounded second moment. The deterministic mean
gradient is bounded on that neighborhood, and hence
$\sup_x\E\norm{\grad\ell(x;Z)}^2<\infty$. The i.i.d. sequence
$\xi_k=\grad\ell(x_\star;Z_k)$ satisfies the functional central limit theorem
and Lindeberg condition under its finite second moment. The remaining
assumptions are imposed in the proposition. Therefore
\Cref{thm:integrated-fclt,thm:selfnorm,cor:functionals} give
\eqref{eq:M-clt} and the confidence procedures in \Cref{prop:M-estimator}.
\end{proof}

\begin{proof}[Proof of \Cref{prop:frechet}]
On a compact set separated from the cut locus, the squared-distance function is smooth jointly in $(x,z)$ and its derivatives are uniformly controlled by the envelope in \Cref{prop:frechet}(a). The gradient identity \eqref{eq:frechet-gradient} follows from the first variation formula. Thus the stochastic gradient at $\mu$ is $-\Log_\mu(Z)$, its mean is zero, and its covariance is $\Sigma_\mu$. Differentiating the mean gradient gives $H_\mu$. The envelope assumptions verify the H\"older Hessian and local $L^2$ noise-continuity conditions.

On a Hadamard manifold, the logarithm is globally single valued and the Fr\'echet mean is unique under the finite squared-distance moment in \Cref{prop:frechet}(b). For bounded support on the affine invariant SPD manifold, compactness of closed geodesic balls and smoothness of squared distance bound its derivatives uniformly on a compact neighborhood of $\mu$ and the support. This implies the local Hessian and noise-continuity bounds. For unbounded support, the local mean-field and noise-continuity conditions remain assumptions in part (b). Once localization and the mean square rate are available, \Cref{thm:selfnorm} gives \eqref{eq:frechet-clt} and the confidence region.
\end{proof}

\begin{proof}[Proof of \Cref{prop:pca}]
The Grassmann tangent space at $[U]$ can be represented by matrices $\Delta$ satisfying $U^\transpose\Delta=0$. Projecting the Euclidean gradient $-XX^\transpose U$ onto this tangent space gives \eqref{eq:pca-gradient}.

Use the eigenbasis $[U_\star,U_\perp]$ and a tangent matrix $K\in\R^{(p-r)\times r}$. A geodesic or any second order local parameterization has expansion
\[
 U(K)=U_\star+U_\perp K-\frac12U_\star K^\transpose K+O(\norm{K}^3).
\]
Substitution into \eqref{eq:pca-objective} gives
\begin{align*}
 F\{[U(K)]\}-F([U_\star])
 &=\frac12\sum_{i=1}^{p-r}\sum_{j=1}^r
 (\lambda_j-\lambda_{r+i})K_{ij}^2
 +O(\norm{K}^3).
\end{align*}
This proves \eqref{eq:pca-H}; the eigengap makes the Hessian positive definite.

At $U_\star$,
\[
 G([U_\star],X)
 =-U_\perp X_\perp X_\parallel^\transpose.
\]
Its expectation is zero because $U_\perp^\transpose CU_\star=0$, and projecting by $U_\perp^\transpose$ before vectorization gives \eqref{eq:pca-Sigma}. Moreover,
$\norm{G([U],X)}\le\norm{X}^2$, so $\E\norm{X}^4<\infty$ gives the gradient second moment. Smoothness of the Grassmann projection implies
\[
 \norm{G([U],X)-\mathsf P_{[U_\star]\to[U]}G([U_\star],X)}
 \le C\norm{X}^2\dist([U],[U_\star])
\]
locally. Squaring and taking expectations verifies \eqref{eq:noise-continuity} with $\nu=1$. The deterministic objective is smooth on the compact Grassmann manifold, so its Hessian is locally Lipschitz. The general theorem now gives \eqref{eq:pca-clt} and the self-normalized region.
\end{proof}


\begin{thebibliography}{99}

\bibitem[Absil et~al.(2008)Absil, Mahony, and Sepulchre]{Absil2008}
Absil, P.-A., Mahony, R., and Sepulchre, R. (2008).
\newblock \emph{Optimization Algorithms on Matrix Manifolds}.
\newblock Princeton University Press.

\bibitem[Afsari(2011)]{Afsari2011}
Afsari, B. (2011).
\newblock Riemannian $L^p$ center of mass: existence, uniqueness, and convexity.
\newblock \emph{Proceedings of the American Mathematical Society}, 139, 655--673.

\bibitem[Balzano et~al.(2010)Balzano, Nowak, and Recht]{Balzano2010}
Balzano, L., Nowak, R., and Recht, B. (2010).
\newblock Online identification and tracking of subspaces from highly incomplete information.
\newblock In \emph{Proceedings of the 48th Annual Allerton Conference on Communication, Control, and Computing}, 704--711.

\bibitem[Bhattacharya and Patrangenaru(2005)]{BhattacharyaPatrangenaru2005}
Bhattacharya, R. and Patrangenaru, V. (2005).
\newblock Large sample theory of intrinsic and extrinsic sample means on manifolds. II.
\newblock \emph{The Annals of Statistics}, 33, 1225--1259.

\bibitem[Bonnabel(2013)]{Bonnabel2013}
Bonnabel, S. (2013).
\newblock Stochastic gradient descent on Riemannian manifolds.
\newblock \emph{IEEE Transactions on Automatic Control}, 58, 2217--2229.

\bibitem[Boumal(2023)]{Boumal2023}
Boumal, N. (2023).
\newblock \emph{An Introduction to Optimization on Smooth Manifolds}.
\newblock Cambridge University Press.

\bibitem[Chen et~al.(2020)Chen, Lee, Tong, and Zhang]{Chen2020}
Chen, X., Lee, J. D., Tong, X. T., and Zhang, Y. (2020).
\newblock Statistical inference for model parameters in stochastic gradient descent.
\newblock \emph{The Annals of Statistics}, 48, 251--273.

\bibitem[Cortez et~al.(2009)Cortez, Cerdeira, Almeida, Matos, and Reis]{Cortez2009}
Cortez, P., Cerdeira, A., Almeida, F., Matos, T., and Reis, J. (2009).
\newblock Modeling wine preferences by data mining from physicochemical properties.
\newblock \emph{Decision Support Systems}, 47, 547--553.

\bibitem[Cortez et~al.(2009)Cortez, Cerdeira, Almeida, Matos, and Reis]{CortezUCI2009}
Cortez, P., Cerdeira, A., Almeida, F., Matos, T., and Reis, J. (2009).
\newblock Wine Quality [Dataset].
\newblock UCI Machine Learning Repository.
\newblock \url{https://doi.org/10.24432/C56S3T}.

\bibitem[Dryden et~al.(2009)Dryden, Koloydenko, and Zhou]{Dryden2009}
Dryden, I. L., Koloydenko, A., and Zhou, D. (2009).
\newblock Non-Euclidean statistics for covariance matrices, with applications to diffusion tensor imaging.
\newblock \emph{The Annals of Applied Statistics}, 3, 1102--1123.

\bibitem[Durmus et~al.(2021)Durmus, Jim\'enez, Moulines, and Said]{Durmus2021}
Durmus, A., Jim\'enez, P., Moulines, E., and Said, S. (2021).
\newblock On Riemannian stochastic approximation schemes with fixed step-size.
\newblock In \emph{Proceedings of the 24th International Conference on Artificial Intelligence and Statistics}, PMLR 130, 1018--1026.

\bibitem[Edelman et~al.(1998)Edelman, Arias, and Smith]{Edelman1998}
Edelman, A., Arias, T. A., and Smith, S. T. (1998).
\newblock The geometry of algorithms with orthogonality constraints.
\newblock \emph{SIAM Journal on Matrix Analysis and Applications}, 20, 303--353.

\bibitem[Eltzner and Huckemann(2019)]{EltznerHuckemann2019}
Eltzner, B. and Huckemann, S. F. (2019).
\newblock A smeary central limit theorem for manifolds with application to high-dimensional spheres.
\newblock \emph{The Annals of Statistics}, 47, 3360--3381.

\bibitem[Fang et~al.(2018)Fang, Xu, and Yang]{FangXuYang2018}
Fang, Y., Xu, J., and Yang, L. (2018).
\newblock Online bootstrap confidence intervals for the stochastic gradient descent estimator.
\newblock \emph{Journal of Machine Learning Research}, 19(78), 1--21.

\bibitem[Fletcher and Joshi(2007)]{FletcherJoshi2007}
Fletcher, P. T. and Joshi, S. (2007).
\newblock Riemannian geometry for the statistical analysis of diffusion tensor data.
\newblock \emph{Signal Processing}, 87, 250--262.

\bibitem[Hall and Heyde(1980)]{HallHeyde1980}
Hall, P. and Heyde, C. C. (1980).
\newblock \emph{Martingale Limit Theory and Its Application}.
\newblock Academic Press.

\bibitem[Hotz et~al.(2024)Hotz, Le, and Wood]{HotzLeWood2024}
Hotz, T., Le, H., and Wood, A. T. A. (2024).
\newblock Central limit theorem for intrinsic Fr\'echet means in smooth compact Riemannian manifolds.
\newblock \emph{Probability Theory and Related Fields}, 189, 1219--1246.

\bibitem[Kurisu and Otsu(2026)]{KurisuOtsu2026}
Kurisu, D. and Otsu, T. (2026).
\newblock Empirical likelihood for manifolds.
\newblock \emph{Journal of the Royal Statistical Society, Series B}, 88, 91--119.

\bibitem[Kushner and Yin(2003)]{KushnerYin2003}
Kushner, H. J. and Yin, G. G. (2003).
\newblock \emph{Stochastic Approximation and Recursive Algorithms and Applications}, second edition.
\newblock Springer.

\bibitem[Lee et~al.(2025)Lee, Liao, Seo, and Shin]{Lee2025QR}
Lee, S., Liao, Y., Seo, M. H., and Shin, Y. (2025).
\newblock Fast inference for quantile regression with tens of millions of observations.
\newblock \emph{Journal of Econometrics}, 249, 105673.

\bibitem[Lee et~al.(2022)Lee, Liao, Seo, and Shin]{Lee2022}
Lee, S., Liao, Y., Seo, M. H., and Shin, Y. (2022).
\newblock Fast and robust online inference with stochastic gradient descent via random scaling.
\newblock \emph{Proceedings of the AAAI Conference on Artificial Intelligence}, 36(7), 7381--7389.

\bibitem[Li et~al.(2022)Li, Liang, Chang, and Zhang]{Li2022}
Li, X., Liang, J., Chang, X., and Zhang, Z. (2022).
\newblock Statistical estimation and online inference via local SGD.
\newblock In \emph{Proceedings of the 35th Conference on Learning Theory}, PMLR 178, 1613--1661.

\bibitem[Mardia and Jupp(2000)]{MardiaJupp2000}
Mardia, K. V. and Jupp, P. E. (2000).
\newblock \emph{Directional Statistics}.
\newblock Wiley.

\bibitem[Oja(1982)]{Oja1982}
Oja, E. (1982).
\newblock Simplified neuron model as a principal component analyzer.
\newblock \emph{Journal of Mathematical Biology}, 15, 267--273.

\bibitem[Pennec(2006)]{Pennec2006}
Pennec, X. (2006).
\newblock Intrinsic statistics on Riemannian manifolds: basic tools for geometric measurements.
\newblock \emph{Journal of Mathematical Imaging and Vision}, 25, 127--154.

\bibitem[Pennec et~al.(2006)Pennec, Fillard, and Ayache]{PennecFillardAyache2006}
Pennec, X., Fillard, P., and Ayache, N. (2006).
\newblock A Riemannian framework for tensor computing.
\newblock \emph{International Journal of Computer Vision}, 66, 41--66.

\bibitem[Polyak and Juditsky(1992)]{PolyakJuditsky1992}
Polyak, B. T. and Juditsky, A. B. (1992).
\newblock Acceleration of stochastic approximation by averaging.
\newblock \emph{SIAM Journal on Control and Optimization}, 30, 838--855.

\bibitem[Shao(2010a)]{Shao2010}
Shao, X. (2010a).
\newblock A self-normalized approach to confidence interval construction in time series.
\newblock \emph{Journal of the Royal Statistical Society, Series B}, 72, 343--366.

\bibitem[Shao(2010b)]{ShaoCorrigendum2010}
Shao, X. (2010b).
\newblock Corrigendum: A self-normalized approach to confidence interval construction in time series.
\newblock \emph{Journal of the Royal Statistical Society, Series B}, 72, 695--696.

\bibitem[Su and Zhu(2023)]{SuZhu2023}
Su, W. J. and Zhu, Y. (2023).
\newblock HiGrad: uncertainty quantification for online learning and stochastic approximation.
\newblock \emph{Journal of Machine Learning Research}, 24(124), 1--53.

\bibitem[Tripuraneni et~al.(2018)Tripuraneni, Flammarion, Bach, and Jordan]{Tripuraneni2018}
Tripuraneni, N., Flammarion, N., Bach, F., and Jordan, M. I. (2018).
\newblock Averaging stochastic gradient descent on Riemannian manifolds.
\newblock In \emph{Proceedings of the 31st Conference on Learning Theory}, PMLR 75, 650--687.

\end{thebibliography}
\end{document}